%% file: paper.tex
\documentclass[letterpaper]{article}
\usepackage[preprint]{aaai2027}
\usepackage[hyphens]{url}
\usepackage{graphicx}
\usepackage{natbib}
\usepackage{caption}
\usepackage{subcaption}
\usepackage{amsmath}
\usepackage{amssymb}
\usepackage[utf8]{inputenc}
\usepackage{booktabs}
\usepackage{tabularx}

\title{Beyond Scene Shortcuts: Action-Centric Video-Text Alignment for Fine-Grained Action Understanding}

\author{Zhiliang Zhu\textsuperscript{1}, Jiazheng Huang\textsuperscript{2}, Dawei Zhao\textsuperscript{1,}\thanks{Corresponding author: \texttt{daweizhao@hrbeu.edu.cn}.}}
\affiliations{
\textsuperscript{1}Harbin Engineering University, China\\
\textsuperscript{2}South China Normal University, China
}

\begin{document}
\maketitle

\begin{figure*}[t]
    \centering
    \begin{subfigure}[b]{0.47\textwidth}
        \includegraphics[width=\textwidth]{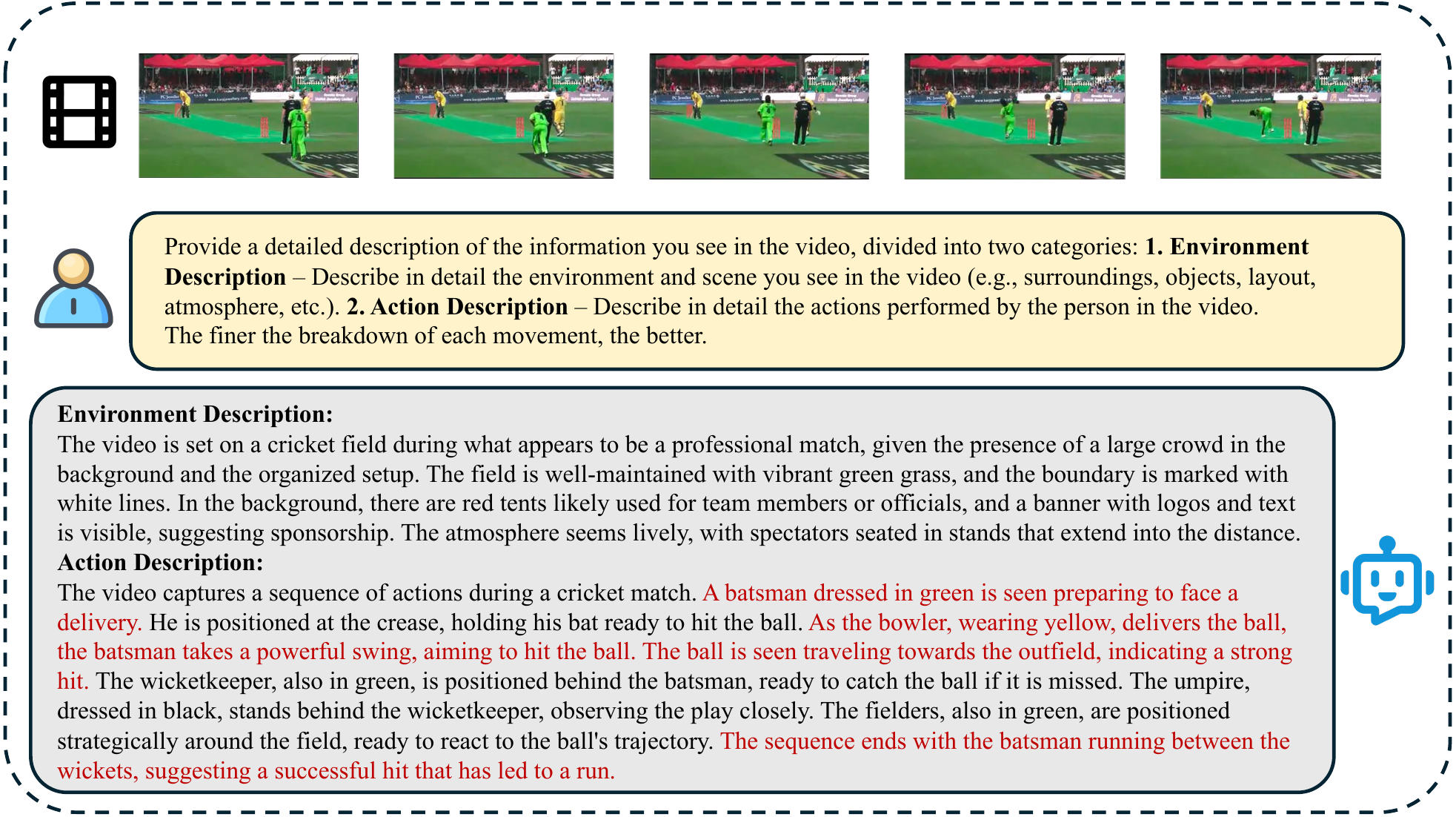}
        \caption{}
        \label{fig:sub2}
    \end{subfigure}
    \begin{subfigure}[b]{0.47\textwidth}
        \includegraphics[width=\textwidth]{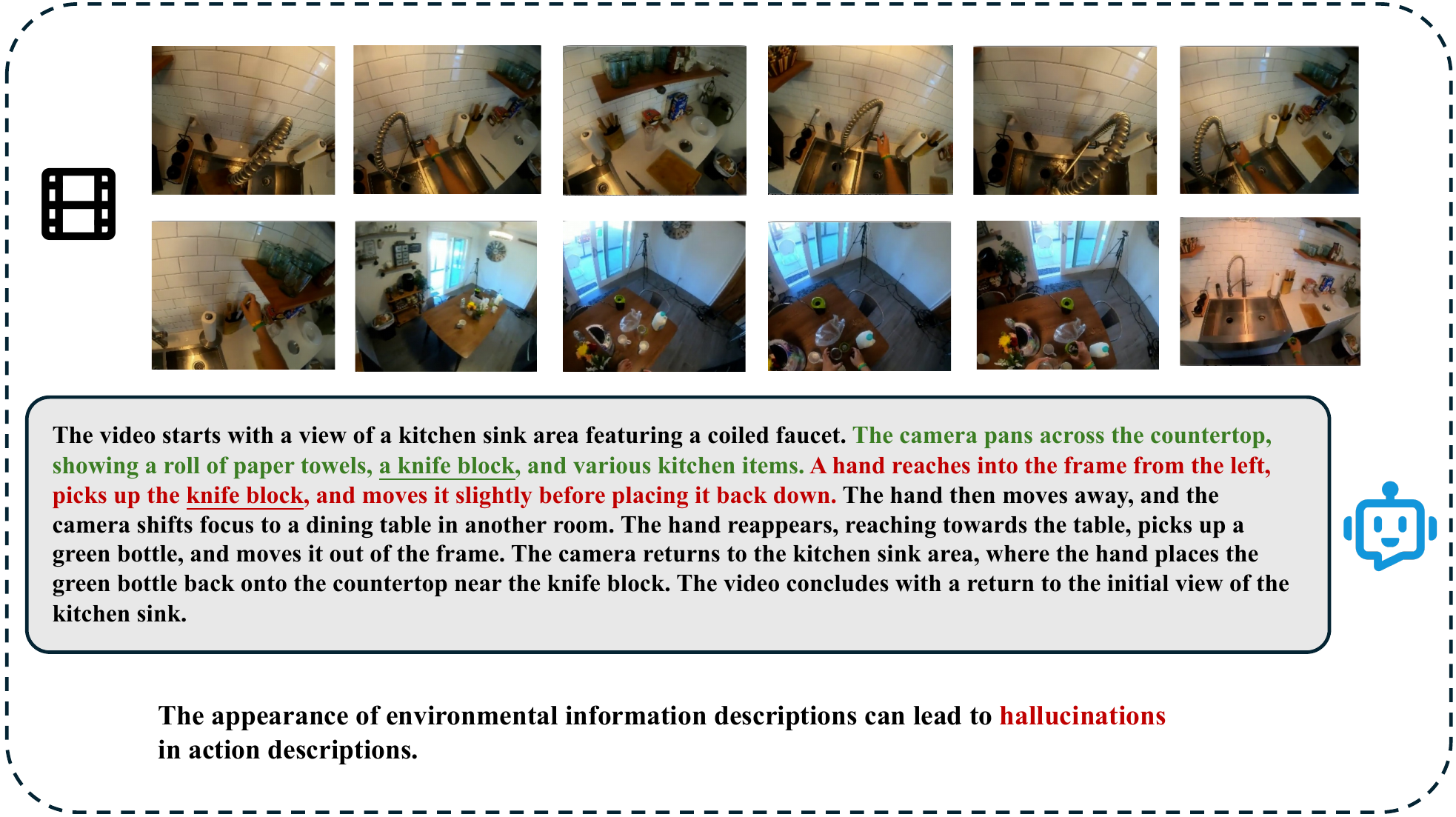}
        \caption{}
        \label{fig:sub3}
    \end{subfigure}
    \caption{
    Illustration of Environment-Grounded Action Hallucination. 
    (a) VLM often generate accurate environment descriptions but incorrect action ones. During SFT, the model develops a strong tendency to anchor action predictions on easily perceived static backgrounds rather than true spatio-temporal evidence, leading to systematically incorrect action descriptions even when its environment descriptions remain accurate. 
    (b) When environmental information is explicitly described in fine-grained action description tasks, it can easily trigger hallucinated actions unrelated to the actual video content.
    }
    \label{fig:before_abstract}
\end{figure*}

\begin{abstract}

Fine-grained action understanding requires vision-language models to ground what an agent does, how it acts, and why it acts in spatio-temporal evidence. However, richer supervision can backfire: fine-tuned models may generate detailed action descriptions grounded in static scenes and salient objects rather than motion, a failure we call Environment-Grounded Action Hallucination (EGAH). To mitigate EGAH, we propose Staged Attention Rectification and Reinforcement (STARR), an action-centric video--text alignment framework that breaks scene--action shortcuts through Action-Environment Dissociation and reinforces action-focused grounding with an Attention Consistency Reward within Group Relative Policy Optimization. We further construct FineActionR\&R, combining temporally localized captions with hierarchical and temporal-causal reasoning data. Built on Qwen2.5-VL-7B, FineActionVLM-R1 improves average MotionBench accuracy by 3.01 percentage points over its base model and achieves 91.50\% on HAU-MCQ-Bench. Qualitative comparisons further indicate more faithful fine-grained action descriptions with fewer environment-induced hallucinations.

\end{abstract}

\section{Introduction}
\label{sec:intro}

Understanding how and why an agent acts---not merely what it does---requires vision-language models (VLMs) to ground action language in subtle spatio-temporal evidence. Fine-grained action understanding therefore depends on motion transitions, body dynamics, interaction order, and object-state changes rather than recognizable scenes and objects alone. For example, saying that ``the robot moves the plate'' misses that it first pauses and then gently slides the plate left to maintain a safe distance from a nearby cup. Despite strong scene and object recognition, recent VLMs still struggle to associate such fine-grained action language with the corresponding dynamic evidence~\cite{qwen2.5-VL,choi2025mashvlmmitigatingactionscene,salah2024enhancingvideotransformers,wu2025hieroactionhierarchicallyguided}.

A natural response is to fine-tune VLMs with richer action captions and reasoning annotations. Yet we observe that this supervision can backfire: fine-tuned models produce longer and more specific descriptions while introducing actions unsupported by the video, especially in out-of-domain (OOD) examples. As illustrated in Figure~\ref{fig:before_abstract}, the environment and salient objects may be described correctly even when the central action is not, suggesting that the model learns where an action is likely to occur rather than identifying the motion that actually occurs. We call this failure Environment-Grounded Action Hallucination (EGAH), a specific form of multimodal hallucination~\cite{zhou2023analyzing,chen2024halc,yang2025mitigating} in which action language is grounded in static environmental cues rather than temporal dynamics.

We attribute EGAH to two coupled biases. First, static scenes and salient objects are visually dominant and persist across frames, making them easier to align with text than short, subtle, or multi-stage movements. A model can therefore reduce its training loss through scene--action correlations instead of learning precise temporal localization. Second, supervised fine-tuning strengthens linguistic co-occurrences between environments and actions, allowing weak visual grounding to expand into a fluent but unsupported narrative. Together, visual shortcuts suggest a plausible action, while linguistic priors elaborate it into a detailed description.

To shift video--text grounding from scene shortcuts to genuine action evidence, we propose Staged Attention Rectification and Reinforcement (STARR), an action-centric video--text alignment framework. Its first component, Action-Environment Dissociation, inserts a temporally annotated action clip into unrelated background sequences while retaining the action caption as the generation target. Because the surrounding environment no longer provides a reliable signal for predicting the caption, the model is encouraged to locate the action-containing segment and ground its description in temporal evidence. Rather than simply adding more supervision, this construction changes which visual evidence is useful for predicting action language.

Even after supervised fine-tuning, action-focused grounding may weaken as autoregressive generation becomes increasingly influenced by linguistic priors. STARR therefore introduces an Attention Consistency Reward within Group Relative Policy Optimization (GRPO)~\cite{shao2024deepseekmath}. The reward promotes strong and balanced attention over action-relevant visual tokens, reinforcing action-centered grounding during generation. The two components are complementary: Action-Environment Dissociation reduces the predictive value of environmental shortcuts, while the Attention Consistency Reward reinforces the use of action-relevant visual evidence.

To connect fine-grained grounding with higher-level action reasoning, we further construct FineActionR\&R, a dataset suite combining temporally localized action captions, hierarchical action-reasoning questions, and temporal-causal multiple-choice questions. Existing video benchmarks largely emphasize action recognition or classification, with limited support for studying how an action unfolds, why a particular strategy is adopted, and how multiple events are temporally or causally related~\cite{hong2025motionbenchbenchmarkingimprovingfinegrained,tu2025favor}. FineActionR\&R provides complementary supervision for recognizing what occurs, describing how it occurs, and reasoning about why it occurs.

FineActionVLM-R1 is trained on FineActionR\&R using the STARR framework, with Qwen2.5-VL-7B as the base model~\cite{qwen2.5-VL}. The resulting model improves average MotionBench accuracy by 3.01 percentage points over its base model and achieves 91.50\% accuracy on HAU-MCQ-Bench. On the latter benchmark, it outperforms substantially larger models, including Qwen2.5-VL-72B and Qwen3-VL-32B, with training exposure differing across models; matched-SFT controls are reported in Appendix~\ref{sec:additional_hau}. Qualitative and out-of-domain analyses further show that FineActionVLM-R1 produces more faithful fine-grained descriptions with fewer environment-induced hallucinations. These results support our central finding: effective fine-grained action understanding requires not only richer supervision, but also explicit control over the visual evidence to which action language is aligned.

Our main contributions are summarized as follows:
\begin{itemize}
    \item We identify and characterize Environment-Grounded Action Hallucination, a fine-grained action grounding failure in which action language relies on static environmental cues rather than motion evidence. We further analyze the roles of visual shortcuts and linguistic priors in producing this behavior.

    \item We propose Staged Attention Rectification and Reinforcement, which combines Action-Environment Dissociation during supervised fine-tuning with an Attention Consistency Reward during GRPO to promote action-centered video--text grounding.

    \item We construct FineActionR\&R, integrating temporally localized action captions with hierarchical and temporal-causal reasoning supervision. Experiments show gains in action recognition and reasoning, while qualitative comparisons indicate more faithful fine-grained captioning.
\end{itemize}


\section{Related Work}

\subsection{Action-aware VLMs}

Recent work on human action understanding has improved recognition robustness, temporal modeling, and fine-grained analysis. MASH-VLM~\cite{choi2025mashvlmmitigatingactionscene} mitigates action--scene hallucination by explicitly disentangling spatio-temporal representations. FTP~\cite{salah2024enhancingvideotransformers} uses VLM-aided training to combine the generalization of VLMs with the specialization of Video Transformers. For finer action distinctions, HieroAction~\cite{wu2025hieroactionhierarchicallyguided} decomposes complex actions into hierarchically organized sub-movements, while FinePseudo~\cite{rizve2024finepseudoimprovingpseudolabelling} exploits temporal alignability to improve pseudo-label quality in semi-supervised recognition. Complementing these model-based advances, MotionBench~\cite{hong2025motionbenchbenchmarkingimprovingfinegrained} and FAVOR-Bench~\cite{tu2025favor} provide dedicated evaluations of fine-grained video motion understanding.

However, this strong body of work remains focused on action recognition—identifying what movement occurred. A key challenge in action understanding lies in inferring the actor’s intent and causal reasoning behind fine- grained actions. We study a supervision-induced generation failure in which rich action descriptions become grounded in static environmental cues. STARR addresses this misalignment at the data and optimization levels, while FineActionR\&R extends supervision from recognizing what occurs to describing how actions unfold and reasoning about their temporal and causal structure.

\subsection{Mitigating Hallucinations in VLMs}

Methods for mitigating VLM hallucinations broadly fall into training-free inference-time interventions and training-based optimization~\cite{chen2024halc,zhou2023analyzing,yang2025mitigating}. Training-free methods modify multimodal decoding without updating model parameters. PAINT~\cite{arif2025paint} amplifies attention to informative visual tokens, IKOD~\cite{yang2025ikod} compresses text tokens through KV-cache merging to increase the relative contribution of visual information, and M3ID~\cite{favero2024multi} contrasts multimodal logits with a text-only prior. Training-based approaches instead strengthen visual grounding through model optimization. Qwen-LA~\cite{chu2025qwen} teaches the model to re-attend to visual tokens during reasoning, while Hallucination-Induced Optimization~\cite{lyu2024alleviating} trains the model to recognize and move away from its self-generated hallucinations.

Most general-purpose hallucination mitigation methods are developed primarily for image-based settings. Fine-grained video action hallucination introduces an additional grounding challenge: the relevant evidence may be brief and distributed across frames, while the surrounding scene can be recognized correctly even when the generated action is unsupported. EGAH captures this specific mismatch between static scene evidence and temporal action language. STARR accordingly combines Action-Environment Dissociation, which reduces the predictive value of environmental shortcuts during supervised fine-tuning, with an Attention Consistency Reward that reinforces action-relevant visual grounding during generation.

\begin{figure*}[t]
    \centering
    \begin{subfigure}[b]{0.47\textwidth}
        \includegraphics[width=\textwidth]{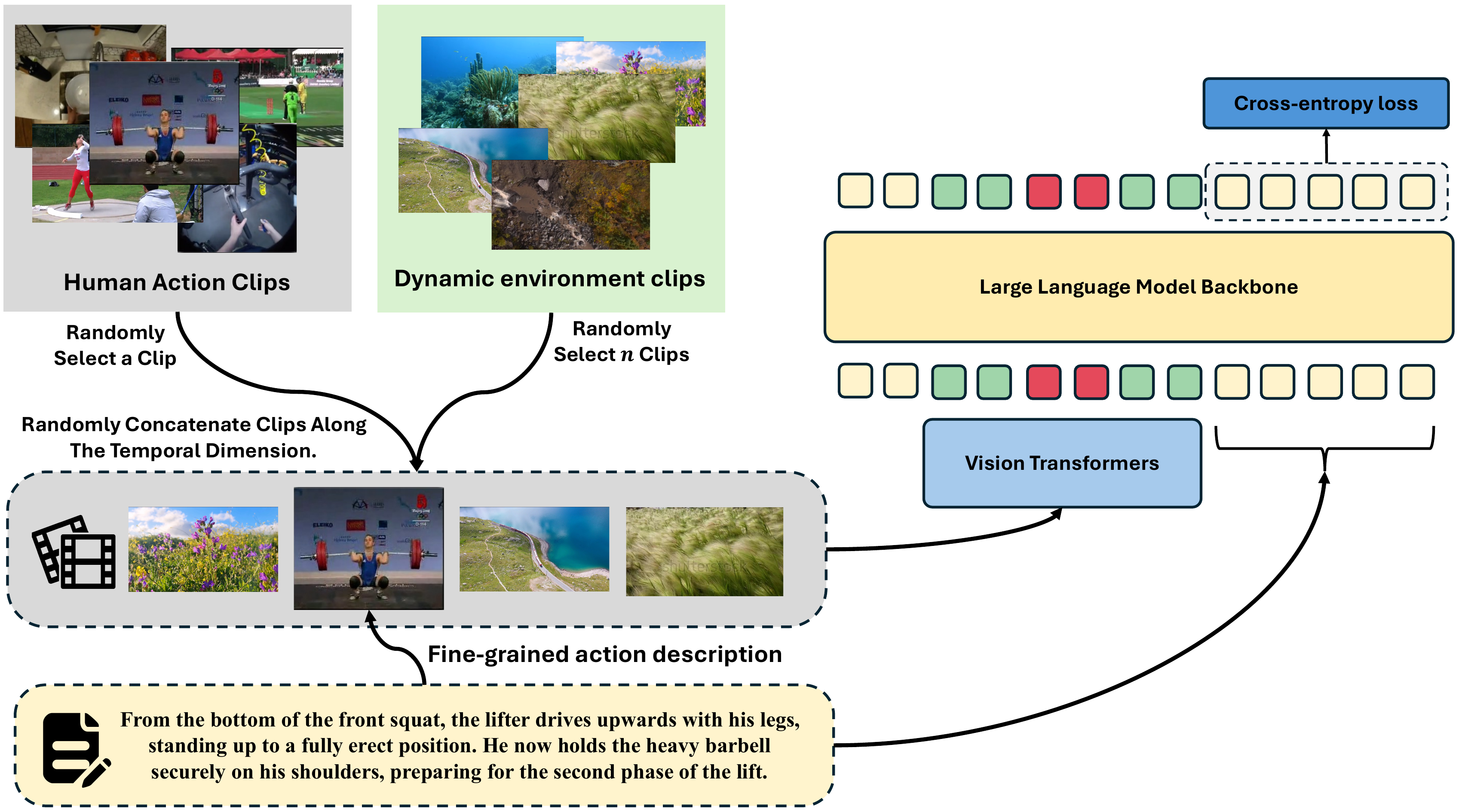}
        \caption{Attention Rectification via Action-Environment Dissociation.}
        \label{fig:SFT}
    \end{subfigure}
    \begin{subfigure}[b]{0.47\textwidth}
        \includegraphics[width=\textwidth]{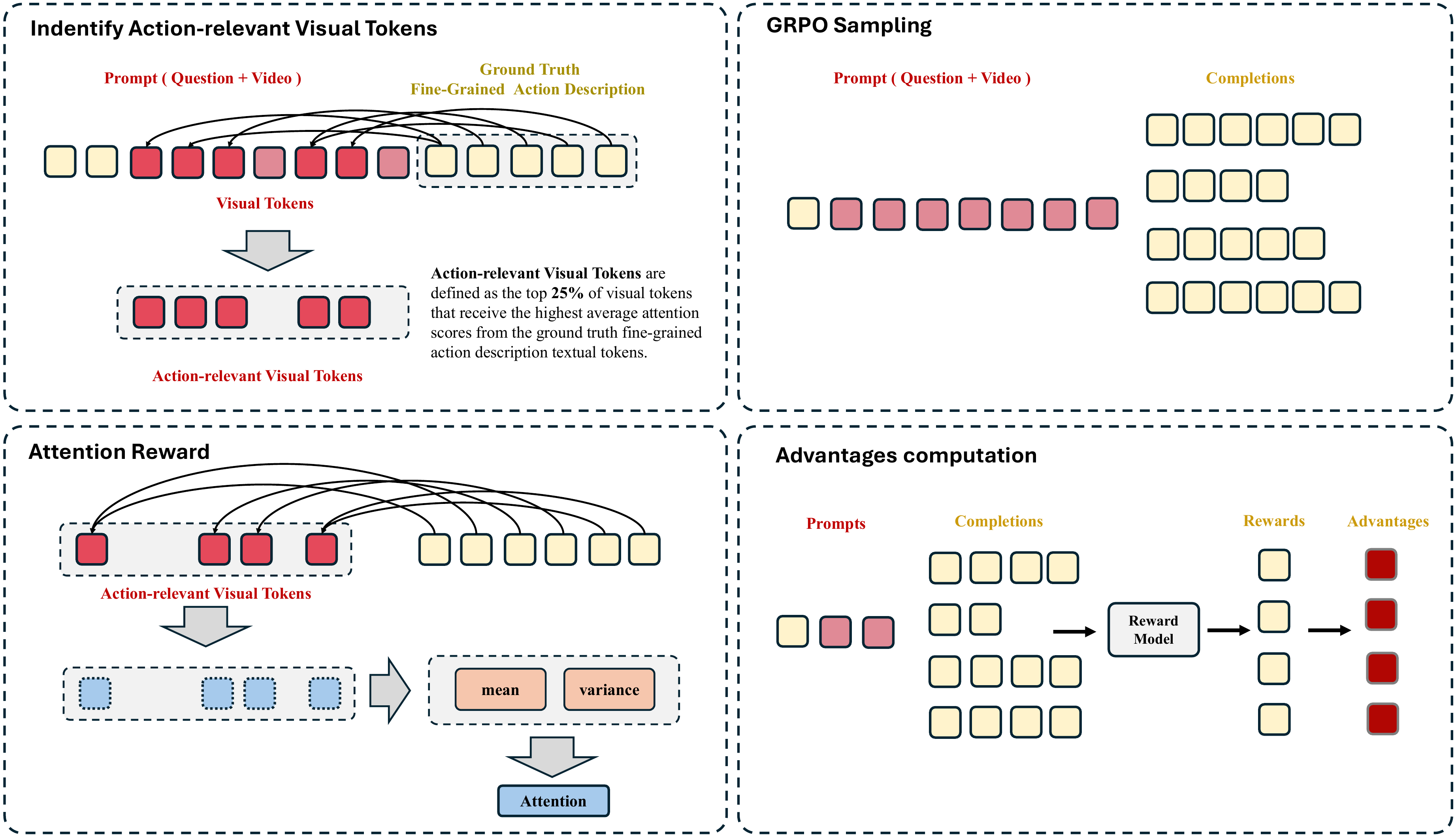}
        \caption{Attention Consistency Reinforcement.}
        \label{fig:GRPO}
    \end{subfigure}
    \caption{
    Staged Attention Rectification and Reinforcement method for fine-grained action understanding. 
    (a) Attention Rectification: SFT is used to train on distractor videos that combine an action clip with irrelevant backgrounds, forcing the model to focus on true action regions. 
    (b) Attention Consistency Reinforcement: GRPO optimizes a reward encouraging stable attention on key action-relevant visual tokens throughout generation.
    }
\end{figure*}


\section{Datasets}
We construct \textbf{FineActionR\&R (Fine-grained Action Recognition \& Reasoning)}, a dataset suite with four components for fine-grained action grounding and reasoning. Additional construction and annotation details are provided in Appendix~\ref{sec:Dataset statistics}.

\textbf{Fine-Action-Caption and Fine-Action-Segment.}
Using videos sampled from established action datasets, Fine-Action-Caption provides dense human-written descriptions that decompose complex actions into ordered sub-events and capture body dynamics, object interactions, and relevant context. Fine-Action-Segment further provides the start and end timestamps of each annotated action.

\textbf{Hierarchical Action Understanding QA (HAU-QA).}
HAU-QA is a large-scale question--answer dataset generated with advanced language models such as Gemini 2.5 Pro. Each QA pair includes a rationale covering micro-level physical interactions, macro-level task logic and causality, and agent--environment interaction.

\textbf{Hierarchical Action Understanding MCQ (HAU-MCQ).}
HAU-MCQ converts HAU-QA into multiple-choice questions with plausible but visually inconsistent distractors, providing discriminative supervision for GRPO-based action reasoning.

\textbf{Temporal Action Sequencing and Causality MCQ (TASC-MCQ).}
TASC-MCQ provides structured video questions about action concurrency, ordering, duration, and causality to improve temporal action recognition and reasoning.

\section{Method}
\subsection{Overview}
To address the problem of Environment-Grounded Action Hallucination in fine-grained action understanding, we propose a two-stage training framework named Staged Attention Rectification and Reinforcement. This framework is designed to systematically decouple the model's dependence on static environments and reinforce its focus on core dynamic content.

\textbf{Stage 1: Attention Rectification.} This stage employs SFT on a specially crafted synthetic dataset. In this dataset, core action video clips are randomly spliced into irrelevant background videos. This forces the model to perform implicit temporal localization to identify the action, thereby learning to ignore environmental distractors and fundamentally rectifying its attention bias.

\textbf{Stage 2: Attention Consistency Reinforcement.} This stage is based on GRPO and introduces a novel Attention Consistency Reward. This reward function incentivizes the model to continuously and stably focus its attention on key action-relevant visual tokens throughout the entire text generation process. This solidifies the corrections from the first stage and further suppresses the negative effects of attention drift and linguistic priors.

\subsection{Attention Rectification via Action-Environment Dissociation. }
\label{section:sft}
To address the model's attention bias towards static environmental cues and the resulting EGAH problem, we propose a SFT stage named ``Attention Rectification''. The core objective of this stage is to compel the model to shift its attention from the static background to the crucial dynamic action content. To achieve this, we devise a data synthesis strategy that eliminates the model's reliance on environmental shortcuts by constructing training samples where the visual environment and the target description are deliberately mismatched.

Specifically, our method involves constructing a synthetic dataset, Action-Environment Dissociation Dataset (AED)  $\mathcal{D}_{syn}$. We first prepare two data sources: a collection of action clips with temporal annotations and corresponding textual descriptions, denoted as $\mathcal{D}_{act} = \{(V_{act}, C_{act})\}$, and a collection of video clips containing only dynamic backgrounds with no significant actions, denoted as $\mathcal{D}_{bg} = \{V_{bg}\}$. For each action clip $V_{act}$ in $\mathcal{D}_{act}$, we randomly sample $n$ background clips $\{V_{bg,i}\}_{i=1}^n$ from $\mathcal{D}_{bg}$. The action clip $V_{act}$ is then inserted at a random temporal position among the sequence of these $n$ background clips, which are then concatenated to form a synthetic video $V_{syn}$. This process yields a new training sample $(V_{syn}, C_{act})$, where the synthetic video $V_{syn}$ serves as the model input and the original action description $C_{act}$ serves as the generation target. During training, for a target caption $C_{act}$ composed of tokens $y_1, \dots, y_T$, the optimization objective for the model $\mathcal{M}$ (with parameters $\theta$) is to minimize the standard autoregressive negative log-likelihood loss (i.e., the next-token prediction loss):

\begin{equation}
    \begin{aligned}
    \mathcal{L}_{\text{SFT}}(\theta)
    = \mathbb{E}_{(V_{\text{syn}}, C_{\text{act}}) \sim \mathcal{D}_{\text{syn}}}
    \bigg[ - \sum_{t=1}^{T} &\log P(y_t \mid V_{\text{syn}}, \\
    &y_{<t}; \theta) \bigg]
    \end{aligned}
\end{equation}

Under this training paradigm, the multiple, randomly concatenated background clips $\{V_{bg,i}\}$ act as strong distractors. Since their visual content is entirely unrelated to the target caption $C_{act}$, any predictive strategy relying on global or static scene features is rendered ineffective. To minimize the loss, the model is forced to perform implicit temporal localization: accurately identifying and attending to the action-containing clip $V_{act}$ within the full video $V_{syn}$, while suppressing interference from the irrelevant background clips. This construction reduces correlations between the target action and surrounding context, encouraging action-focused alignment while retaining the scene and object cues within the action clip.


\subsection{Attention Consistency Reinforcement}
\label{section:ACR}
Following attention rectification in the SFT stage, \textbf{Attention Consistency Reinforcement} uses GRPO with an \textbf{Attention Consistency Reward} (ACR). This reward encourages strong and balanced caption-averaged attention to reference-selected visual evidence. Its variance term operates across visual tokens after averaging over generated text positions; it does not directly measure stability across decoding steps. Token-selection validation and component ablations are provided in Appendix~\ref{sec:additional_evaluation}.

To compute the ACR, we must first define a set of action-relevant visual tokens. This is achieved by leveraging the ground-truth fine-grained action caption associated with each video. To generate a stable attention baseline, we use the model's policy from immediately before the current optimization step as a frozen reference model. We then compute the attention from the entire caption to the full set of visual tokens, $F_V = {f_1, \dots, f_N}$. For each visual token, its cumulative attention score is calculated by summing the attention it receives from all textual tokens in the caption. Based on these scores, all visual tokens are ranked. The top 25\% of this ranking are designated as the \textbf{action-relevant visual tokens}, $F_{\text{act}} \subset F_V$. These tokens constitute the critical visual evidence the model is rewarded for focusing on.

To capture the model's attention behavior, we define the attention matrix $A \in \mathbb{R}^{|F_{\text{act}}| \times T}$, where each element $A_{ij}$ denotes the attention score that the $j$-th generated text token $y_j$ allocates to the $i$-th action-relevant visual token $f_i \in F_{\text{act}}$. For each such token, we compute its average attention value across the $T$ generated text tokens:
\begin{equation}
    a_i = \frac{1}{T} \sum_{j=1}^{T} A_{ij} \,.
\end{equation}
The set $\{a_i \,|\, f_i \in F_{\text{act}}\}$ thus characterizes the model's attention distribution within the action-relevant visual space.

\textbf{Maximizing Mean Attention ($R_{\text{mean}}$):}
    This component encourages strong overall attention to action-relevant regions. However, relying solely on an absolute measure can be insufficient when different generated samples yield numerically close mean attention scores. To create a more robust reward signal that differentiates well even in such cases, we formulate $R_{\text{mean}}$ as a composite of two sub-rewards: an absolute reward $R_{\text{abs}}$ and a relative reward $R_{\text{rel}}$. The absolute reward $R_{\text{abs}}$ measures the raw attention strength against a predefined threshold $\tau_m$, rewarding the model for achieving a high baseline of attention. It is defined as:
    \begin{equation}
        R_{\text{abs}} = \min\!\left( \frac{1}{\tau_m |F_{\text{act}}|} \sum_{f_i \in F_{\text{act}}} a_i,\; 1.0 \right).
    \end{equation}
The relative reward $R_{\text{rel}}$ amplifies the differences between samples within the same generation batch. Let $\bar{a}_k = \frac{1}{|F_{\text{act}}|} \sum_{f_i \in F_{\text{act}}} a_i^{(k)}$ be the mean attention for the $k$-th sample in a batch $B$. The relative reward for this sample is calculated by applying min-max normalization across the batch, which highlights its performance relative to its peers:
\begin{equation}
    R_{\text{rel}}^{(k)} = \frac{\bar{a}_k - \min_{j \in B}(\bar{a}_j)}{\max_{j \in B}(\bar{a}_j) - \min_{j \in B}(\bar{a}_j) + \epsilon},
\end{equation}
where $\epsilon$ is a small constant to ensure numerical stability. The final mean attention reward, $R_{\text{mean}}$, is a weighted sum of these two components, allowing us to simultaneously encourage high absolute attention and reward the best-performing samples within a group:
\begin{equation}
    R_{\text{mean}} = \alpha_{\text{abs}} R_{\text{abs}} + \alpha_{\text{rel}} R_{\text{rel}},
\end{equation}
where $\alpha_{\text{abs}}$ and $\alpha_{\text{rel}}$ are hyperparameters balancing the two objectives.

\textbf{Minimizing Attention Variance ($R_{\text{var}}$):}
    This term promotes balanced attention across all action-relevant regions. It rewards lower variance in $\{a_i\}$, ensuring that attention is not concentrated on a specific sub-region. The reward attains its maximum (1.0) when attention is uniformly distributed and decreases linearly as the variance approaches a threshold $\tau_v$:
    \begin{equation}
        R_{\text{var}} = \max\!\left( 0,\; 1 - \frac{\text{Var}_{f_i \in F_{\text{act}}}(a_i)}{\tau_v} \right).
    \end{equation}

Finally, the total \textbf{ACR} combines these two components:
\begin{equation}
    R_{\text{ACR}} = \lambda_{\text{mean}} R_{\text{mean}} + \lambda_{\text{var}} R_{\text{var}} \,,
\end{equation}
where $\lambda_{\text{mean}}$ and $\lambda_{\text{var}}$ are weighting coefficients that control the trade-off between overall attention strength and spatial uniformity.

Intuitively, $R_{\text{mean}}$ encourages attention strength over the reference-selected tokens, while $R_{\text{var}}$ encourages balanced coverage of those tokens after averaging over generated positions. Integrating $R_{\text{ACR}}$ into GRPO directly optimizes this attention objective. Its contribution to caption faithfulness is evaluated through the controlled ablations in Appendix~\ref{sec:additional_ablation}.

\subsection{Training Recipe}
\label{sec:training_recipe}
Our training methodology employs a three-stage paradigm to progressively impart the model with action understanding and description capabilities. This staged approach systematically addresses the challenges of attention bias and linguistic overfitting. The entire training process begins with \textbf{Qwen2.5-VL-7B-Instruct}\cite{qwen2.5-VL} as the base model.

\paragraph{Stage 1: Temporal Reasoning Pre-training}
The initial stage aims to build a foundational capacity for temporal reasoning and causality analysis. Starting with base model, we utilize the \textbf{TASC-MCQ} within the GRPO. This process encourages the model to generate intermediate reasoning steps before providing a final answer, effectively stimulating its ability to reason about the temporal dynamics presented in a video, rather than relying on static priors. The outcome of this stage is an initial model checkpoint, \textbf{FineActionVLM-R1-Zero}, which demonstrates a nascent ability for video-grounded thinking.

\paragraph{Stage 2: Attention Rectification and Multi-task Joint Fine-tuning}
This stage is designed to tackle the core issue of EGAH while simultaneously enhancing the model's fine-grained action recognition and understanding abilities. We perform a multi-task SFT on the base \textbf{Qwen2.5-VL-7B-Instruct} model, integrating several key datasets:
(1) The synthetic AED dataset is used to rectify the model's attention bias, as described above. (2) The \textbf{Fine-Action-Caption} dataset and a subset of the \textbf{HAU-QA} dataset are incorporated to strengthen the model's ability to recognize and comprehend nuanced human actions. (3) A curated dataset for video-grounded reasoning is also included. This data is created by distilling reasoning chains from the \textbf{FineActionVLM-R1-Zero} model on the \textbf{TASC-MCQ}, followed by LLM-based filtering to remove samples reliant on common-sense rather than video observation. Through this comprehensive SFT process, we obtain the \textbf{FineActionVLM-R1-Dev1} model. This model shows significant improvement in mitigating EGAH and is capable of generating higher-quality samples for both in-domain and out-of-domain data.

\paragraph{Stage 3: Attention Consistency Reinforcement and Comprehensive Capability Enhancement}
Starting from the \textbf{FineActionVLM-R1-Dev1} checkpoint, this final stage aims to stabilize generation quality and fully activate its advanced reasoning capabilities using a GRPO-based reinforcement learning paradigm. We introduce the novel ACR with the \textbf{Fine-Action-Caption} dataset. This reward mechanism reinforces the strength and balance of caption-averaged attention over reference-selected visual tokens. Grounded in the principle that action recognition and understanding are mutually reinforcing, we continue to incorporate the \textbf{TASC-MCQ} and \textbf{HAU-MCQ} datasets. Training on these discriminative multiple-choice tasks further refines the model's ability to discern subtle temporal and causal relationships. This final stage yields our ultimate model, \textbf{FineActionVLM-R1}.

\begin{table*}[t]
    \centering 
    \small
    \setlength{\tabcolsep}{1pt}
    \begin{tabularx}{\textwidth}{@{}c>{\raggedright\arraybackslash}X>{\centering\arraybackslash}p{0.75in}cc*{7}{c}@{}}
        \toprule
        \textbf{\#} & \textbf{Model} & \textbf{Organization} & \textbf{Frames} & \shortstack{\textbf{LLM}\\\textbf{Params}} & \shortstack{\textbf{Dev}\\\textbf{Avg (\%)}} & \textbf{MR (\%)} & \textbf{LM (\%)} & \textbf{CM (\%)} & \textbf{MO (\%)} & \textbf{AO (\%)} & \textbf{RC (\%)} \\
        \midrule

        1 & Qwen3-VL-30B-A3B-Instruct & Alibaba & 2fps & 30B (A3B) & 57.59 & 61.03 & 61.90 & 55.84 & 75.07 & 45.09 & 26.75 \\
        2 & Qwen3-VL-8B-Instruct & Alibaba & 2fps & 8B & 55.55 & 59.07 & 57.88 & 55.84 & 72.46 & 44.70 & 24.00 \\
        3 & Kimi-VL-2506 & Kimi & 2fps & 16B(A3B) & 54.6 & 58.7 & 55.78 & 45.64 & 73.48 & 42.08 & 22.52 \\
        4 & Qwen2.5-VL-72B-Instruct & Alibaba & 1fps & 72B & 57.00 & 58.00 & 61.00 & \underline{63.00} & 72.00 & 47.00 & 31.00 \\
        5 & gemini-2.5-flash & Google & 2fps & - & 56.12 & 59.25 & 59.97 & 59.9 & 72.55 & 44.4 & 29.14 \\
        6 & GLM-4.5V & Zhipu AI & 2fps & 106B(A12B) & \underline{61.42} & \textbf{65.08} & 65.16 & 58.72 & \textbf{77.38} & \underline{48.34} & \underline{40.37} \\
        7 & GLM-4V-Plus-0111 & Zhipu AI & 2fps & 106B(A12B) & \textbf{62.8} & \underline{64.1} & \textbf{67} & \textbf{67.4} & 73.5 & 46.7 & \textbf{42.8} \\
        8 & GLM-4.1V-Thinking & Zhipu AI & 2fps & 9B & 58.68 & 62.48 & 62.98 & 53.85 & \underline{76.18} & 46.23 & 33.51 \\
        9 & gpt-5-chat & OpenAI & 2fps & - & 58.81 & 61.37	& \underline{66.67}	& 58.61	& 73.79	& \textbf{49.38}	& 26.74 \\
        10 &Qwen2.5-VL-7B-Instruct (Base) & Alibaba & 2fps & 7B & 54.83 & 58.25 & 57.69 & 49.09 & 72.90 & 38.92 & 33.25 \\
        11 & \textbf{FineActionVLM-R1 (ours)} & - & 2fps & 7B & 57.84 & 62.92 & 60.81 & 51.94 & 73.77 & 43.74 & 31.5 \\
        \bottomrule
    \end{tabularx}
    \caption{Performance comparison of different models on \textbf{MotionBench}, which evaluates fine-grained action perception across six core tasks: Motion Recognition (MR), Location-related Motion (LM), Camera Motion (CM), Motion-related Objects (MO), Action Order (AO), and Repetition Count (RC). Dev Avg is overall accuracy across questions, weighted by task size (Appendix~\ref{sec:additional_protocol}).}
    \label{tab:experiment_results}
\end{table*}

\begin{table*}[htbp]
    \centering
    \small
    \setlength{\tabcolsep}{2pt}
    \begin{tabularx}{\textwidth}{@{}>{\raggedright\arraybackslash}Xccc@{}}
        \toprule
        \textbf{Model} & \shortstack{\textbf{Avg.}\\\textbf{Accuracy(\%)}} & \shortstack{\textbf{Seen Videos +}\\\textbf{Unseen Questions.(\%)}} & \shortstack{\textbf{Unseen Videos +}\\\textbf{Unseen Questions.(\%)}} \\
        \midrule
        Qwen3-VL-8B-Instruct & 63.75 & 60.95 & 66.95 \\
        Qwen3-VL-8B-Thinking & 67.2  & 65.92  & 68.66  \\
        Qwen3-VL-32B-Instruct & \underline{78.75}  & \underline{76.12}  & \underline{81.77}  \\
        Qwen3-VL-32B-Thinking & 75.43  & 72.64  & 78.63  \\
        Qwen3-VL-30B-A3B-Instruct & 72.11  & 71.39  & 72.93  \\
        Qwen3-VL-30b-A3B-Thinking & 71.31  & 72.93  & 69.9  \\
        GLM-4.5V & 73.97  & 73.88  & 74.07  \\
        GLM-4V-plus-0111 & 65.07  & 66.67  & 63.25  \\
        Qwen2.5-VL-7B-Instruct & 45.29  & 45.27  & 45.3  \\
        Qwen2.5-VL-72B-Instruct & 64.01  & 59.2  & 69.52  \\
        \textbf{FineActionVLM-R1(ours)} & \textbf{91.50}  & \textbf{92.54}  & \textbf{90.31}  \\
        \bottomrule
    \end{tabularx}
    \caption{Performance comparison of different models on \textbf{HAU-MCQ-bench}. All models were evaluated at a frame rate of 2\,fps.}
    \label{tab:HAU-MCQ-Bench}
    
\end{table*}

\section{Experiment}
In this section, we conduct comprehensive experiments to evaluate both the effectiveness of our FineActionR\&R dataset and the proposed STARR method. Specifically, we first demonstrate that training on FineActionR\&R significantly improves the model's capability in fine-grained action recognition and understanding, supported by results on MotionBench and HAU-MCQ-Bench. These findings verify the dataset's strong contribution to enhancing temporal perception and hierarchical reasoning. Furthermore, to validate the effectiveness of our STARR method, we perform caption quality assessments, showing that it substantially alleviates EGAH and enables the generation of temporally consistent, visually grounded fine-grained descriptions. Additional qualitative analyses are presented in Appendix~\ref{sec:Supplementary Experiment}; human evaluation, component ablations, and quantitative controls are reported in Appendix~\ref{sec:additional_evaluation}.

\subsection{Benchmarks}
\textbf{Fine-grained Action Recognition.} We use MotionBench\cite{hong2025motionbenchbenchmarkingimprovingfinegrained} to evaluate our model's fine-grained action perception. The benchmark assesses a model's perception of subtle, transient dynamics across six core tasks: Motion Recognition, Location-related Motion, Action Order, Repetition Count, Motion-related Objects, and Camera Motion.

\textbf{Fine-grained Action Understanding.} Current public video understanding benchmarks primarily focus on action recognition and classification, while generally lacking assessments for the deep underlying reasons (e.g., intent, motivation) behind fine-grained actions. To fill this gap, we have constructed a new evaluation benchmark, named HAU-MCQ-bench, specifically designed to measure a model's reasoning capabilities in this area. This benchmark is derived from our self-constructed HAU-MCQ dataset. We extracted 753 MCQs from it to serve as the test set (HAU-MCQ-Bench), with the remaining portion forming the training set (HAU-MCQ-Train). To evaluate the model's generalization ability, we have carefully partitioned HAU-MCQ-Bench into two distinct evaluation settings:

\textbf{(1) Seen Videos, Unseen Questions:} This split, accounting for 50\% of the benchmark, involves questions whose corresponding videos have appeared in the HAU-MCQ-Train. The model's comprehension of these videos may be enhanced by its exposure to other related questions during training.

\textbf{(2) Unseen Videos, Unseen Questions:} The other half (50\%) consists of questions whose corresponding videos are entirely new and do not appear in the HAU-MCQ-Train.

\begin{figure}[t]
    \centering
    \begin{subfigure}[b]{0.45\textwidth}
        \includegraphics[width=\textwidth]{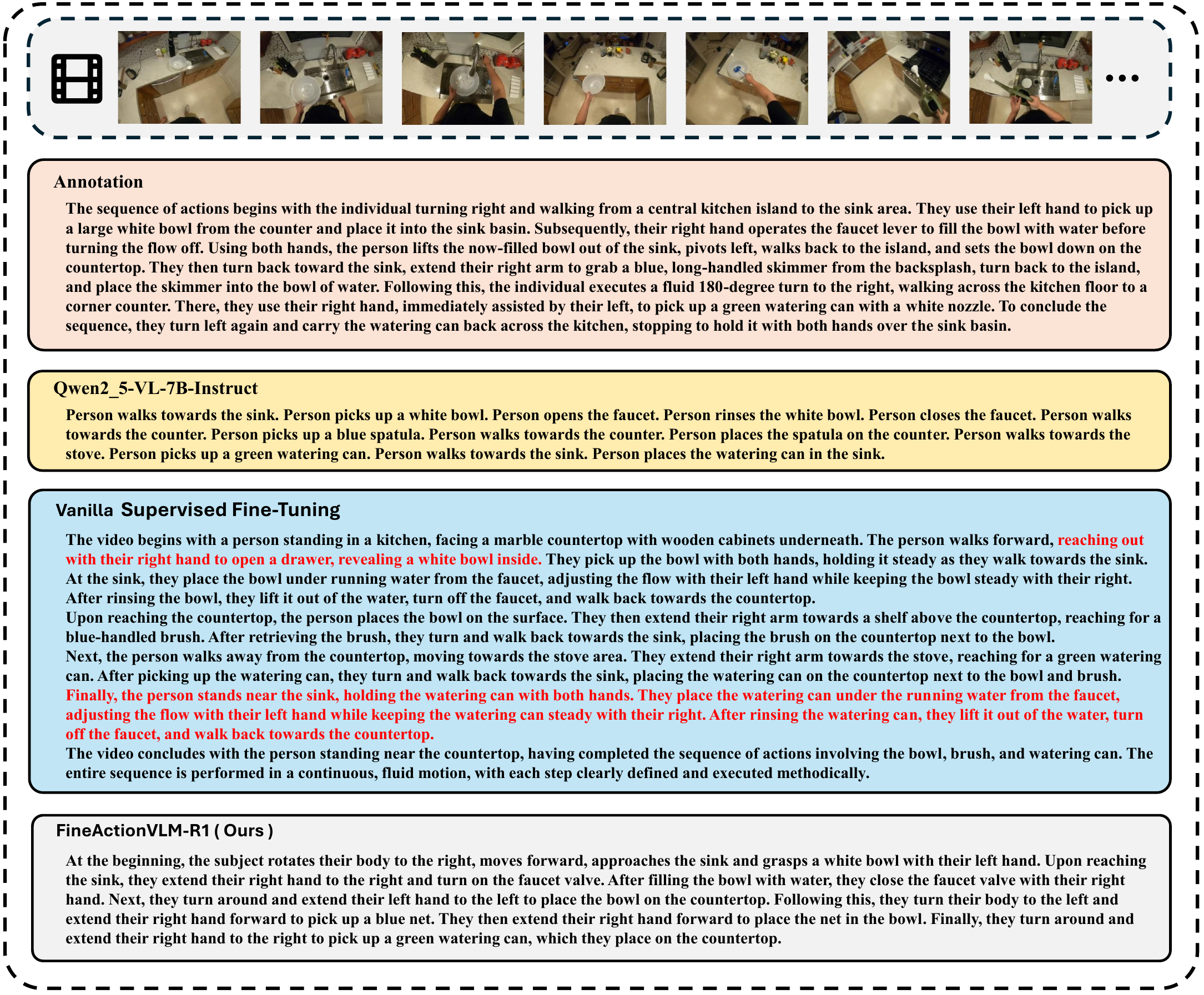}
        \caption{In domain example 1.}
    \end{subfigure}
    \begin{subfigure}[b]{0.45\textwidth}
        \includegraphics[width=\textwidth]{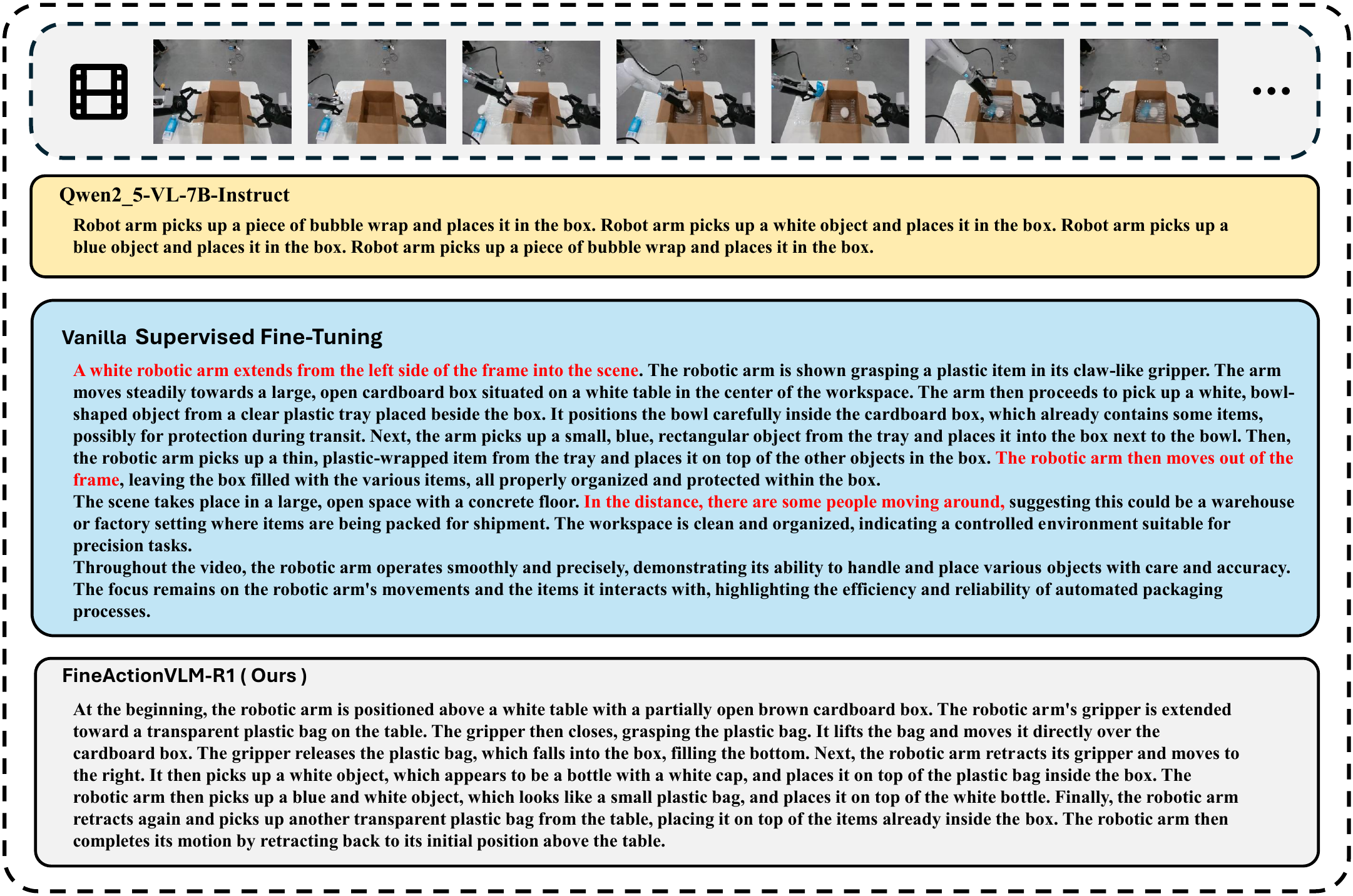}
        \caption{Out of domain example 1.}
    \end{subfigure}
    \caption{Caption Quality Assessment}
    \label{fig:experiment_cap}
\end{figure}

\subsection{MotionBench}
As shown in Table~\ref{tab:experiment_results}, our model, FineActionVLM-R1, achieves an average accuracy of \textbf{57.84\%} on MotionBench, marking an improvement of \textbf{3.01 percentage points} over its base model, Qwen2.5-VL-7B-Instruct (54.83\%). This result indicates that our proposed STARR training paradigm not only effectively mitigates hallucinations but also substantially enhances the model's fundamental perception of underlying dynamic information. Notably, despite its 7B parameter scale, our model outperforms larger or comparable models such as Qwen3-VL-8B-Instruct (55.55\%) and Kimi-VL (54.6\%), and demonstrates competitive performance against leading closed-source models like gpt-5-chat (58.81\%).

A detailed analysis of the sub-task scores reveals that the most significant gains are in \textbf{MR} and \textbf{AO}, the two tasks most relevant to core dynamic understanding. Our model achieves scores of \textbf{62.92\% (+4.67\%)} and \textbf{43.74\% (+4.82\%)} in these areas, respectively. This strongly validates the success of our method in decoupling actions from their environment and reinforcing attention on temporal sequences. Although there is a slight decrease in the RC task (-1.75\%), the overall performance boost confirms the effectiveness and efficiency of the STARR paradigm in enhancing fine-grained action perception.

\subsection{Caption Quality Assessment}
We compare the base model, \textbf{vanilla SFT}, and \textbf{FineActionVLM-R1} under identical prompts on in-domain and OOD videos (Figure~\ref{fig:experiment_cap}). Base produces brief, coarse descriptions, while vanilla SFT adds detail but may infer unsupported actions from scene cues, such as opening a drawer and retrieving a bowl. R1 better captures action transitions and procedural detail in these examples. Further examples appear in Appendix~\ref{sec:Caption Quality Supplementary Assessment}. Direct human evaluation and controlled ablations in Appendix~\ref{sec:additional_ablation} quantify caption faithfulness and distinguish the contributions of AED and ACR.

\subsection{HAU-MCQ-Bench}
We use our custom HAU-MCQ-Bench to evaluate the model's reasoning ability regarding the underlying causes of actions. As shown in Table~\ref{tab:HAU-MCQ-Bench}, our FineActionVLM-R1 achieves an average accuracy of \textbf{91.50\%} on this benchmark, which is higher than all compared baselines. Compared with its base model, Qwen2.5-VL-7B-Instruct (45.29\%), it attains an absolute improvement of \textbf{46.21\%}. It also performs better than larger-scale models, such as Qwen3-VL-32B-Instruct (78.75\%) and Qwen2.5-VL-72B-Instruct (64.01\%), indicating that our training strategy effectively enhances fine-grained action understanding without relying on model scale.

These quantitative results strongly indicate that our training paradigm has successfully endowed and activated the model's ability to think about and understand fine-grained actions. The consistently high performance in both the ``Seen Videos, Unseen Questions'' (\textbf{92.54\%}) and the more challenging ``Unseen Videos, Unseen Questions'' (\textbf{90.31\%}) settings further demonstrates the generalizability of this acquired ability.

\section{Conclusions and Limitations}
This work identifies Environment-Grounded Action Hallucination (EGAH) in fine-grained action understanding, where action language is spuriously grounded in static environmental cues rather than motion evidence. To mitigate EGAH, we propose STARR, a two-stage framework that redirects attention toward dynamic action cues and reinforces action-centered grounding. We further construct FineActionR\&R to support fine-grained action recognition and hierarchical reasoning. Together, these components improve action recognition, reasoning, and caption faithfulness. In the supplemental human evaluation, Full reduces AH from 28.76\% to 19.71\% and raises action recall from 70.23\% to 75.55\% relative to matched SFT. The ablations show an ACR benefit without ES, while sampling controls and additional benchmarks support the gains across the evaluated settings (Appendix~\ref{sec:additional_evaluation}).

However, our method and model possess several limitations that highlight important avenues for future research.

\textbf{Implicit Environmental Bias in Visual Tokens.}
Attention Rectification reduces explicit environment--action correlations but cannot eliminate implicit visual biases. Even without environmental text, salient static-object tokens may dominate attention after projection into the language space, triggering object-related action hallucinations.

\textbf{Limited Generalizability to Out-of-Domain Data.}
Attention Consistency Reinforcement remains less robust on OOD videos, where diffuse attention may overlook critical action segments and yield incomplete or incoherent descriptions.

\textbf{Dependence on Data Diversity and Scale.}
Performance remains dependent on training-data diversity and scale. Novel actions or scenes may lead to coarse descriptions or plausible but unsupported details driven by linguistic priors.
\clearpage
\bibliography{main}

\clearpage
\appendix
\setcounter{secnumdepth}{2}
\twocolumn[\section*{Appendix}]

\section{FineActionR\&R Dataset Statistics}
\label{sec:Dataset statistics}

To comprehensively evaluate both fine-grained human action understanding and temporal-causal reasoning, we construct the \textbf{FineActionR\&R} dataset suite. It consists of two complementary components:
\begin{itemize}
    \item \textbf{(1) Fine-grained Human Action} --- focusing on detailed action perception, segmentation, and basic reasoning.
    \item \textbf{(2) Temporal Action Sequencing and Causality Multiple-Choice Question (TASC-MCQ)} --- targeting higher-level temporal and causal reasoning over complex action sequences.
\end{itemize}
Together, these two parts establish a diverse and scalable benchmark for evaluating models' multi-level understanding of dynamic human actions.

\subsection{Fine-grained Human Action}

To build the fine-grained human action dataset suite, which includes \textit{Fine-Action-Caption}, \textit{Fine-Action-Segment}, \textit{HAU-QA}, and \textit{HAU-MCQ}, we carefully sampled video clips from a wide range of mainstream public video datasets. The selected sources encompass videos from \textbf{FineAction}~\cite{liu2022fineaction} (162 videos), \textbf{UCF101}~\cite{soomro2012ucf101} (255 videos), \textbf{Movid}~\cite{chen2024motionllm} (175 videos), \textbf{Egotaskqa}~\cite{jia2022egotaskqa} (133 videos), \textbf{Oops}~\cite{epstein2019oops} (76 videos), and \textbf{Something-Something V2}~\cite{goyal2017something}, forming a foundational pool of \textbf{857} unique videos.

This diverse mixture ensures a balanced coverage across action scales and viewpoints: \textbf{UCF101} contributes standard third-person actions; \textbf{Egotaskqa} focuses on egocentric, task-oriented interactions; \textbf{Something-Something V2} emphasizes fine-grained, object-centric manipulations; while \textbf{FineAction}, \textbf{Movid}, and \textbf{Oops} add complex, long-horizon, and unexpected behaviors.

Building on this video base, we derived the following datasets with distinct annotation methodologies: 
\begin{itemize}
    \item \textit{Fine-Action-Segment}: fully manually annotated temporal segments;
    \item \textit{HAU-QA} and \textit{HAU-MCQ}: automatically generated with Gemini 2.5 Pro, followed by manual spot-checking for correctness and quality.
\end{itemize}

\begin{figure}[t]
    \centering
    \begin{subfigure}[b]{0.25\textwidth}
        \includegraphics[width=\textwidth]{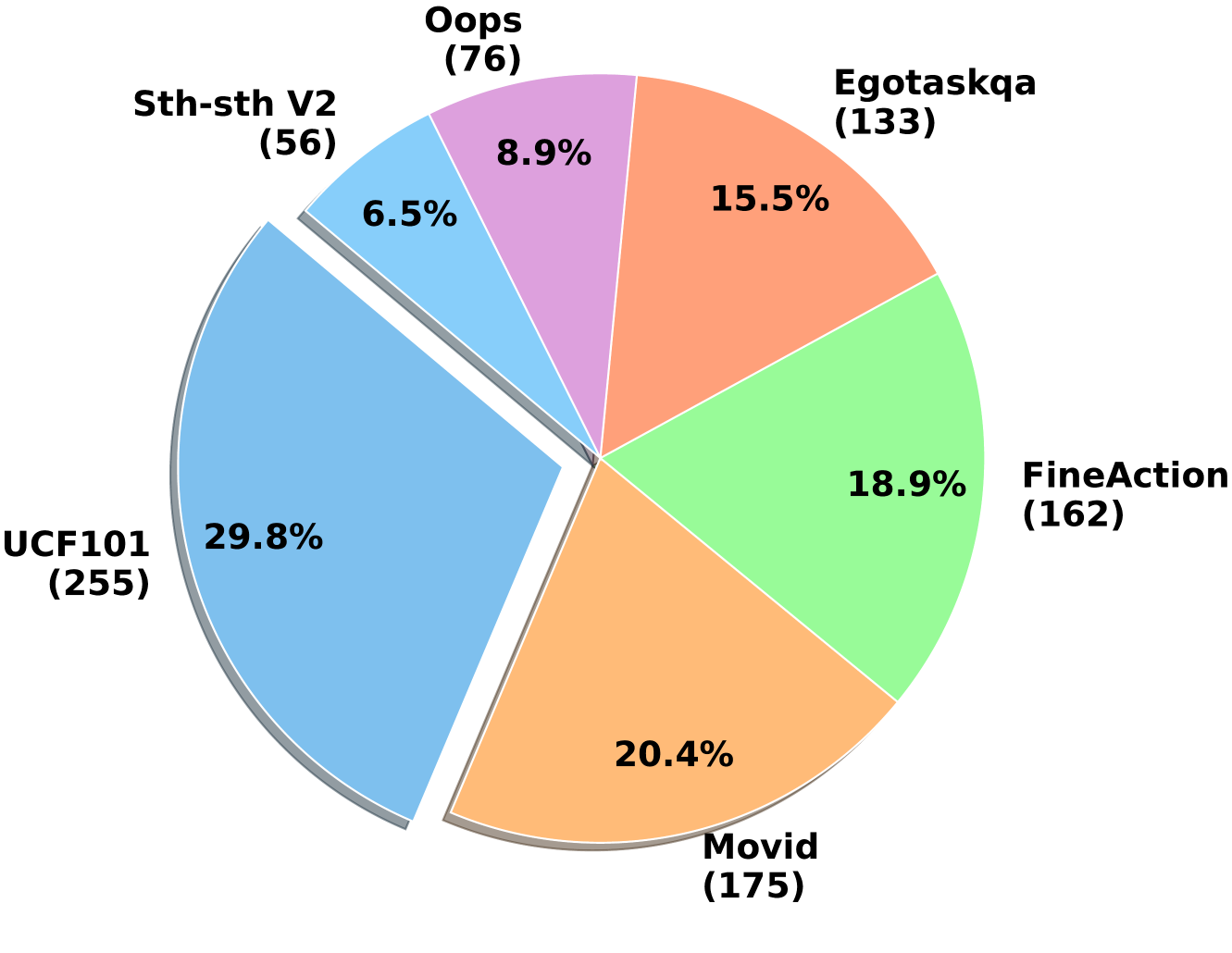}
        \caption{Video source distribution across six datasets.}
        \label{fig:Video Dataset Distribution}
    \end{subfigure}
    \begin{subfigure}[b]{0.20\textwidth}
        \includegraphics[width=\textwidth]{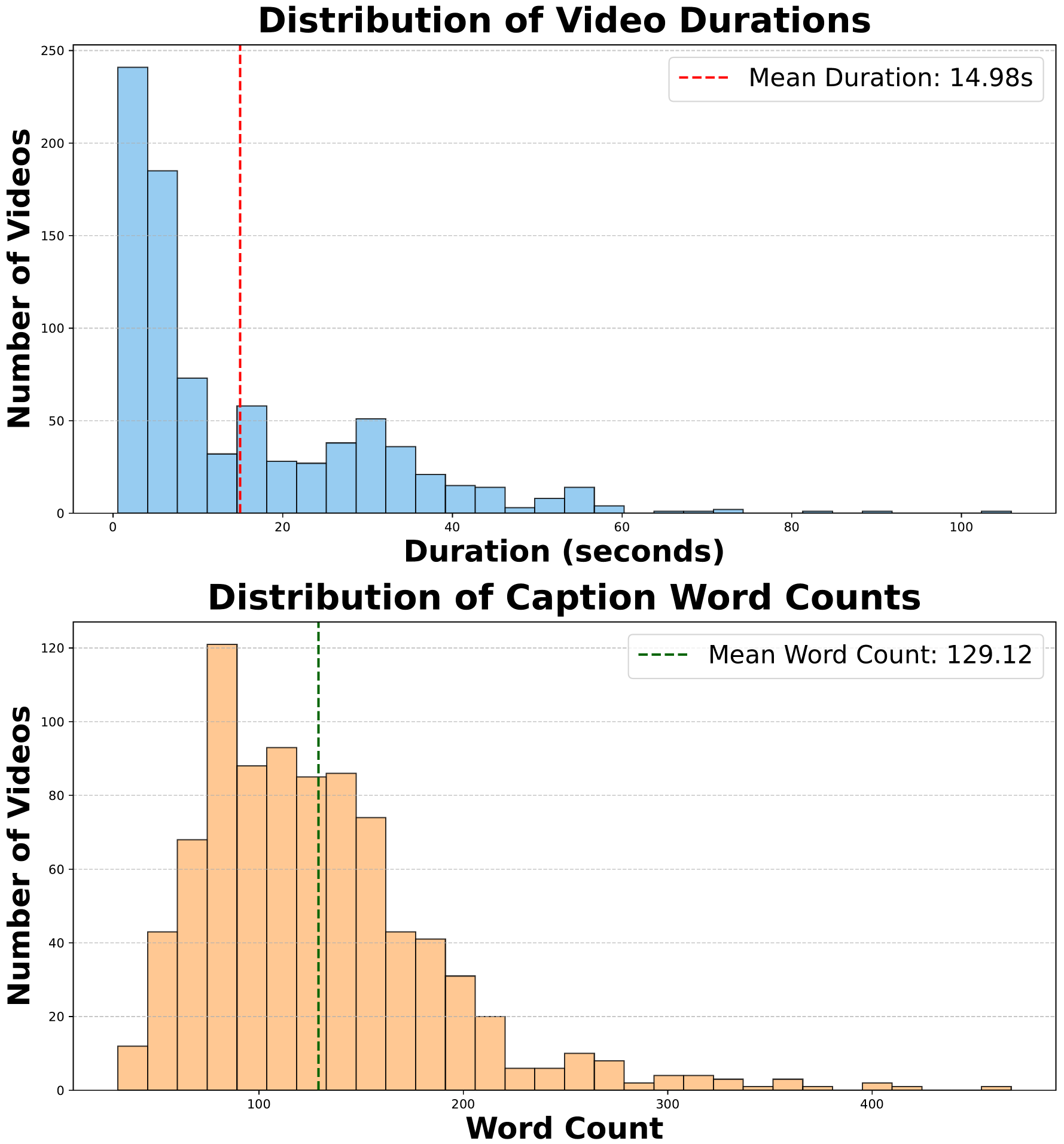}
        \caption{Distributions of video duration and caption length.}
    \end{subfigure}
    
    \vspace{0.3cm} 
    \begin{subfigure}[b]{0.45\textwidth}
        \includegraphics[width=\textwidth]{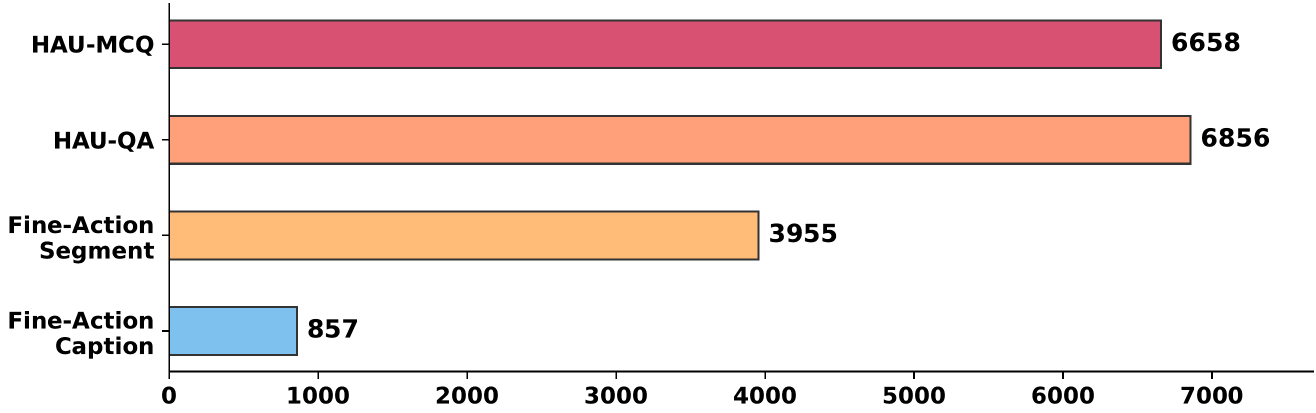}
        \caption{Sample counts of each Fine-grained Human Action subset.}
    \end{subfigure}
    \caption{Statistics of the \textbf{Fine-grained Human Action} dataset. 
    (a)~Video source distribution across six public datasets, showing balanced diversity in action types and visual domains. 
    (b)~Distributions of video durations and caption word counts, indicating that clips are generally short and accompanied by concise textual descriptions. 
    (c)~Counts of data samples across four derived subsets---\textit{Fine-Action-Caption}, \textit{Fine-Action-Segment}, \textit{HAU-QA}, and \textit{HAU-MCQ}---demonstrating the overall scale of the dataset suite.}
    \label{fig:FineAction-statistics}
\end{figure}

\begin{figure*}[p]
    \centering
    \includegraphics[
        width=0.94\textwidth,
        height=0.80\textheight,
        keepaspectratio
    ]{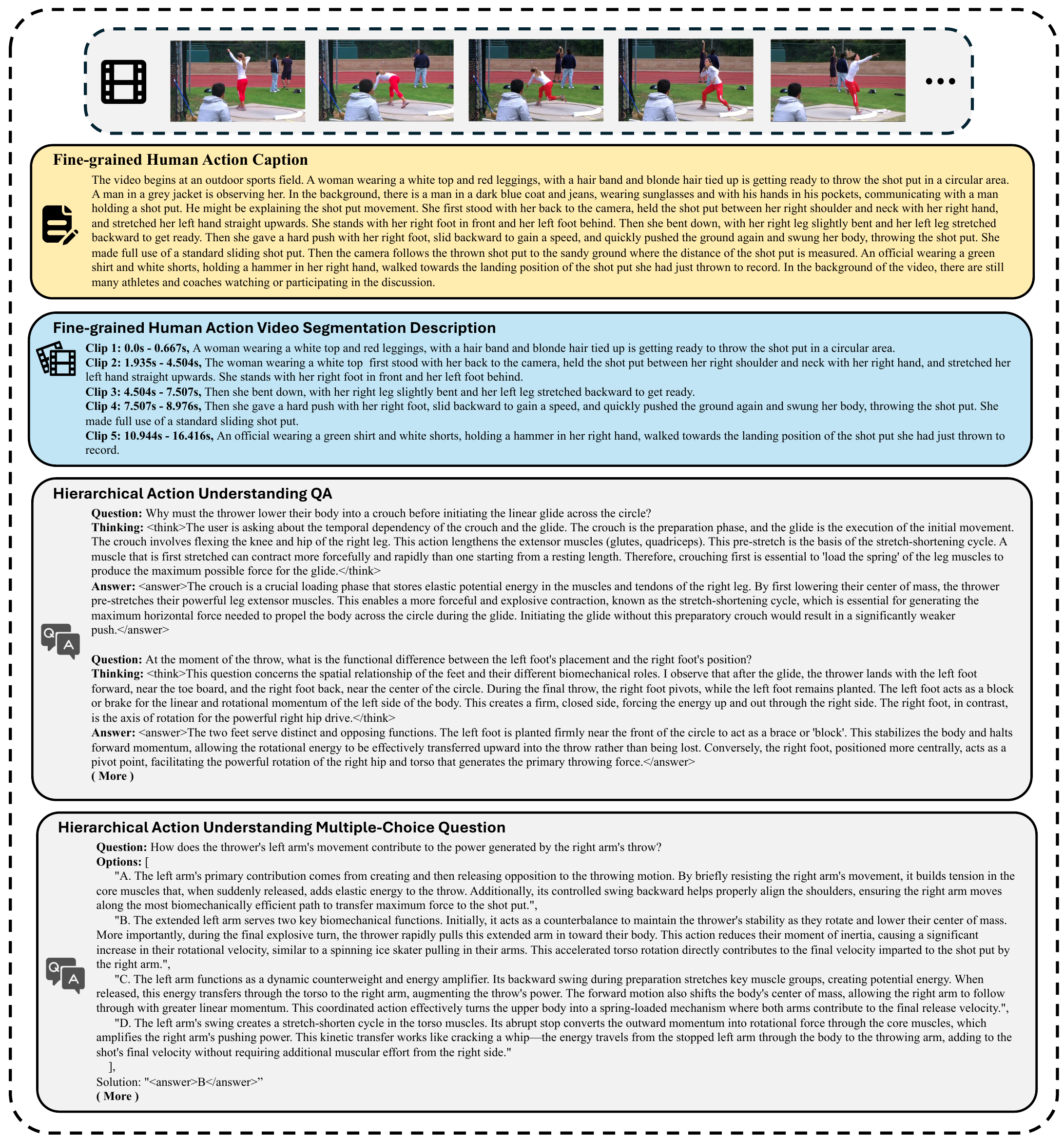}
    \caption{An example annotation from the Fine-grained Human Action dataset.}
    \label{fig:Dataset Demo}
\end{figure*}

Figure~\ref{fig:FineAction-statistics} illustrates the dataset composition and annotation statistics, while Figure~\ref{fig:Dataset Demo} presents a qualitative annotation example.

\FloatBarrier


\subsection{Temporal Action Sequencing and Causality Multiple-Choice Question Dataset}

The second component of FineActionR\&R is the \textbf{Temporal Action Sequencing and Causality Multiple-Choice Question} (TASC-MCQ) dataset, designed to evaluate a model's reasoning capability over temporal and causal relationships in video sequences. It was constructed using a combination of clips from the \textbf{Fine-Action-Caption} and \textbf{FAVOR-Bench}~\cite{tu2025favor} datasets, totaling \textbf{1,948} videos.

Using Gemini-2.5-Pro\cite{comanici2025gemini}, we generated multiple-choice questions covering five aspects of temporal-causal reasoning, and each question features a unique correct answer and semantically-relevant distractors that serve as hard negatives, driving the model to learn fine-grained visual discrimination.:
\begin{itemize}
    \item \textbf{Concurrent Actions:} Identifying simultaneous actions performed by one or more subjects;
    \item \textbf{Local Sequence:} Determining an action immediately preceding or following a key event;
    \item \textbf{Action Duration \& Transition:} Assessing durations or transitions between actions;
    \item \textbf{Global Ordering:} Arranging three or more non-adjacent actions in the correct order;
    \item \textbf{Causality:} Identifying an action that directly triggers a later outcome.
\end{itemize}

\begin{figure}[t]
    \centering
    \begin{subfigure}[b]{0.35\textwidth}
        \includegraphics[width=\textwidth]{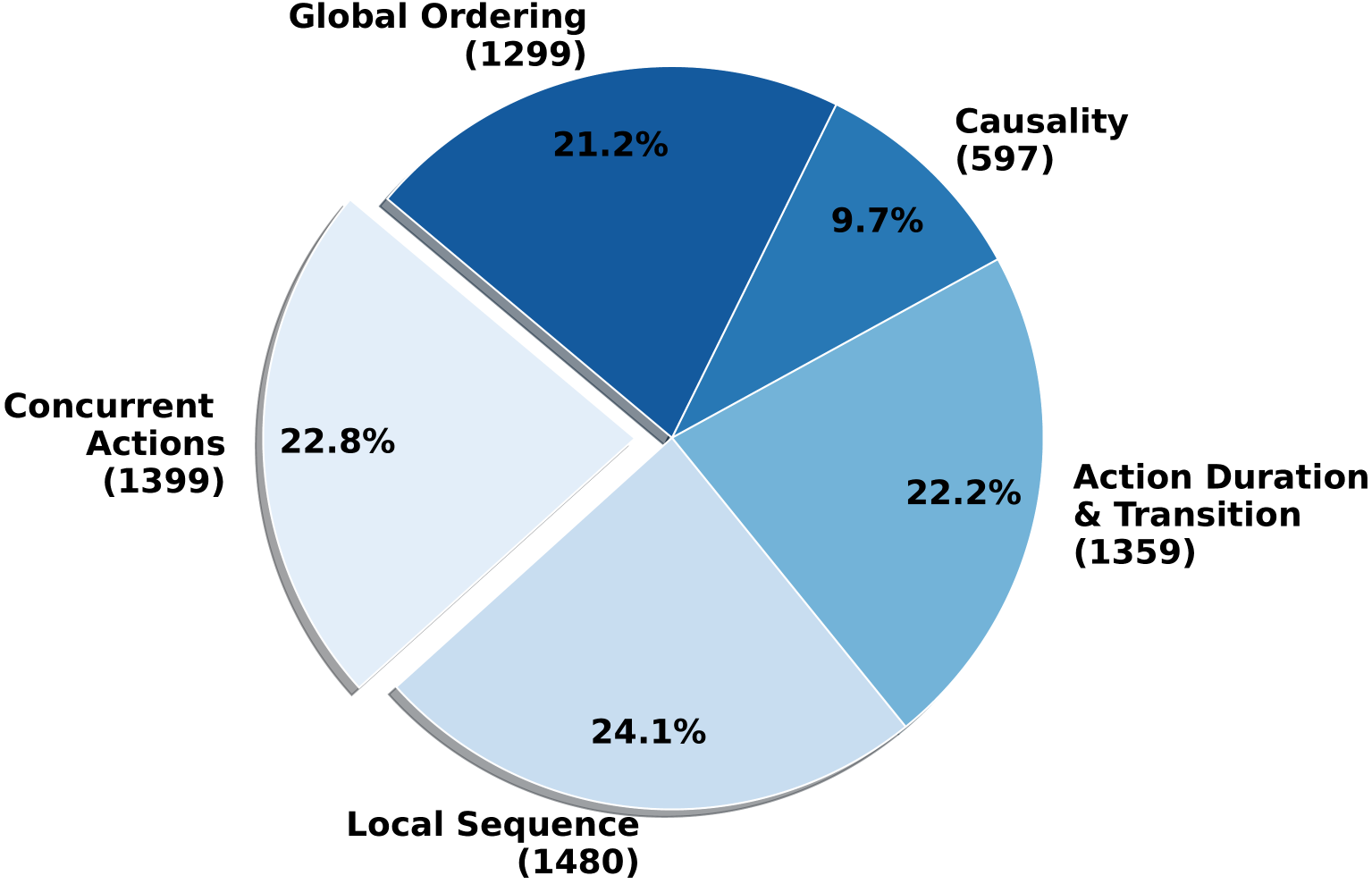}
        \caption{Distribution of question types in TASC-MCQ.}
    \end{subfigure}
    \begin{subfigure}[b]{0.35\textwidth}
        \includegraphics[width=\textwidth]{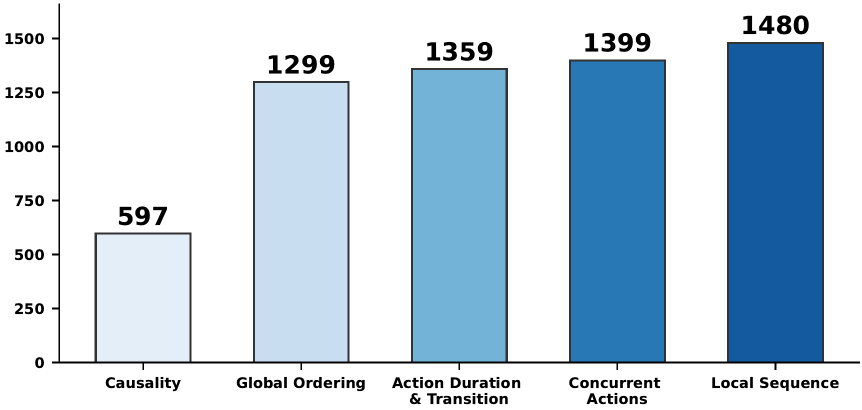}
        \caption{Sample counts for each question type.}
    \end{subfigure}
    \caption{Statistics of the \textbf{TASC-MCQ} dataset. 
    (a)~Distribution of the five question types, including Concurrent Actions, Local Sequence, Action Duration \& Transition, Global Ordering, and Causality. 
    (b)~Counts of samples within each question type, showing a balanced coverage across diverse temporal and causal reasoning categories.}
    \label{fig:TASC-MCQ-statistics}
\end{figure}

To ensure data quality, the generated questions underwent a two-stage refinement process. First, we employed DeepSeek-R1~\cite{guo2025deepseek} to filter out linguistically-answerable questions, ensuring genuine visuo-temporal reasoning requirements. Then, manual spot-checking was performed to verify correctness. The final dataset contains \textbf{6,134} high-quality MCQs focused on fine-grained temporal reasoning. Figure~\ref{fig:TASC-MCQ-statistics} summarizes the distribution of different question types.

\begin{figure}[t]
    \centering
    \includegraphics[width = 8 cm]{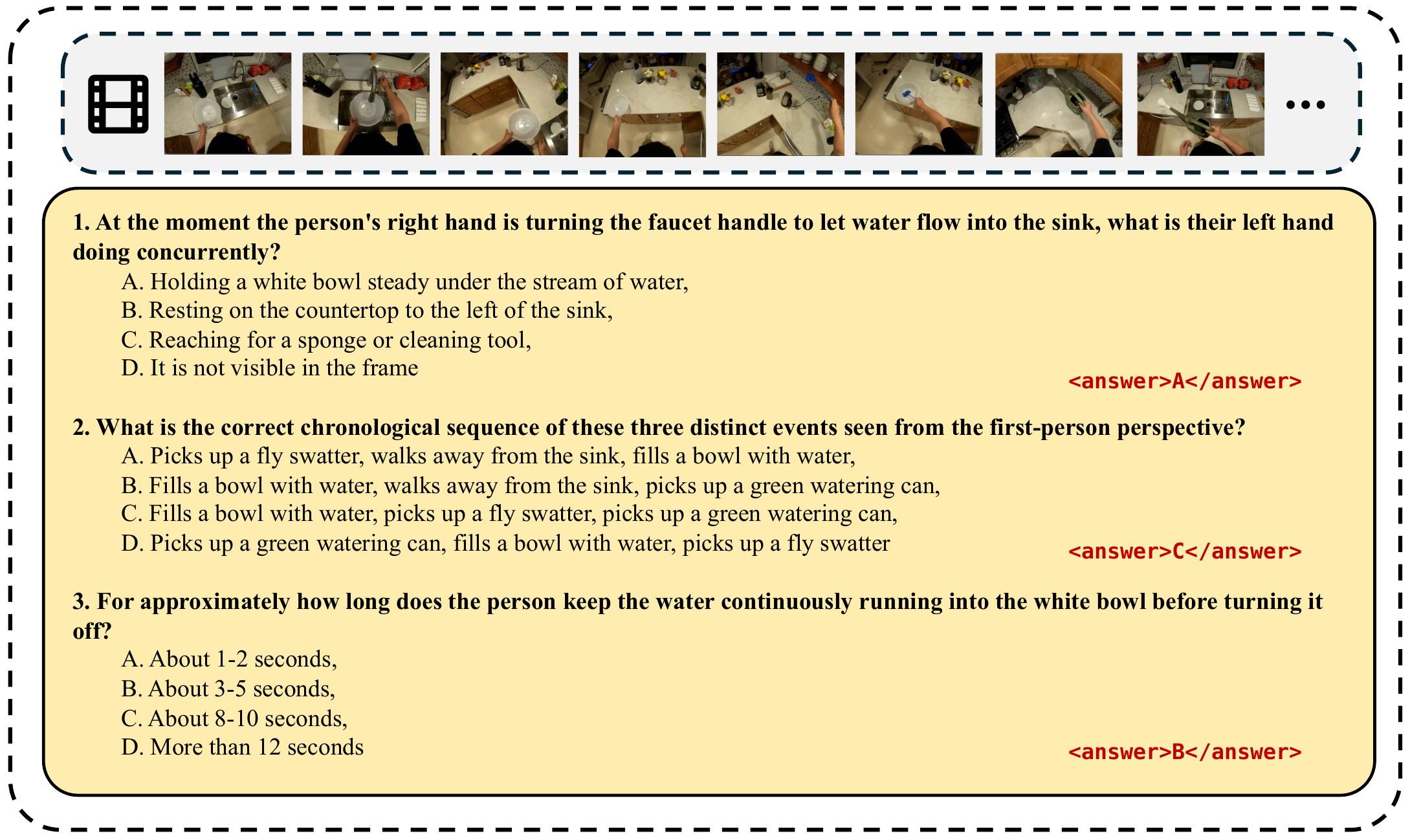}
    \caption{Temporal Action Sequencing and Causality MCQ.}
    \label{fig:Temporal MCQ}
\end{figure}

\section{Training Detail}
\label{sec:appendix_training}
\subsection{Hyperparameters}
In this SFT phase, the model is trained for two epochs using a constant learning rate of $1 \times 10^{-6}$. For video data preprocessing, we sample frames at a rate of 2 FPS. To handle inputs of varying resolutions uniformly, we cap the maximum number of pixels per video frame at 100,352.

In GRPO stage largely inherits the training configuration from the SFT phase. Specifically, the learning rate is maintained at $1 \times 10^{-6}$, and the video processing parameters, including the sampling rate (2 FPS) and the maximum pixel count (100,352), remain identical. During the sampling step of the GRPO process, we generate four candidate samples for each input prompt using a sampling temperature of 1.0. A core component of this stage is a custom \textbf{Attention Consistency Reward} function. This reward is formulated as a weighted sum of a mean consistency term and a variance consistency term, with their respective coefficients set to $\lambda_{\text{mean}}=0.7$ and $\lambda_{\text{var}}=0.3$. Furthermore, the reward calculation balances an absolute component and a relative component, governed by the coefficients $\alpha_{\text{abs}}=0.7$ and $\alpha_{\text{rel}}=0.3$, respectively.

\subsection{GRPO Reward Function}
\label{sec:grpo_reward}

In the GRPO training stage, we designed a composite reward function to synergistically optimize the model's generation behavior across multiple dimensions, particularly for fine-grained action understanding and description tasks. This function is composed of the following core components, each targeting one or more specific objectives to collectively guide the model toward the desired convergence.

\begin{enumerate}
    \item \textbf{Attention Consistency Reward:} ACR is applied to the fine-grained action captioning task. It rewards strong mean attention and balanced coverage across reference-selected visual tokens after averaging over generated text positions. It does not directly measure attention stability across decoding steps. Appendix~\ref{sec:additional_tokens} validates the token selection and clarifies this distinction.

    \item \textbf{Accuracy Reward:} This reward ensures the faithfulness and correctness of the generated content. Its form is contingent on the task type:
    \begin{itemize}
        \item For Multiple-Choice Questions (HAU-MCQ, TASC-MCQ): We provide a fixed positive reward if the model's selected answer option (e.g., ``A'', ``B'') matches the ground truth, and zero reward otherwise. This provides direct binary feedback on correctness.
        \item For the action captioning task (Fine-Action-Caption): We calculate the ROUGE score between the generated caption and the ground-truth caption, using the score as the reward value. This serves to maintain the factual accuracy of the generated content and prevents the model from deviating excessively from the visual facts while optimizing for other rewards like ACR.
    \end{itemize}

    \item \textbf{Formatting Reward:} We employ a reward based on regular expressions to enforce specific output formats required by different tasks. This is crucial for guiding the model to perform structured reasoning and provide structured answers.
    \begin{itemize}
        \item For Multiple-Choice Questions: The output is required to strictly adhere to the \texttt{<think>...}\allowbreak\texttt{</think>}\allowbreak\texttt{<answer>...}\allowbreak\texttt{</answer>} structure. This format compels the model to first generate a Chain-of-Thought (CoT) before providing the final answer, thereby explicitly stimulating its reasoning process.
        \item For the action captioning task: The output format is expected to be \texttt{<answer>...</answer>}.
    \end{itemize}

    \item \textbf{Length Reward:} This reward encourages the model to generate text of an appropriate length, avoiding outputs that are either too long or too short. Its implementation also varies by task:
    \begin{itemize}
        \item For Multiple-Choice Questions: We reward the length of the Chain-of-Thought within the \texttt{<think>} block for falling within a predefined reasonable range. This encourages sufficient but not overly verbose reasoning.
        \item For the action captioning task: We reward the length of the text within the \texttt{<answer>} block. \textbf{This design is critical as it directly counteracts a potential side effect of optimizing ACR alone.} We observed that without this reward, ACR would incentivize the model to generate shorter captions (as shorter text makes it easier to maintain high attention consistency). However, this leads to coarser-grained descriptions that omit fine-grained dynamic details. The length reward thus ensures that the model can generate content-rich, detailed descriptions while maintaining a focused attention.
    \end{itemize}

    \item \textbf{Environmental Information Suppression Reward:} To consolidate the gains from the SFT stage in mitigating EGAH, we introduce this reward. We observed that the inherent ``attention sink'' problem in Vision-Language Models can cause some visual tokens identified by ACR as ``action-relevant'' to actually correspond to static objects in the background. Without suppression, the model might re-introduce descriptions of the environment during the RL stage, causing the EGAH problem to resurface. To address this, we employ a powerful external Large Language Model as a judge to evaluate the content generated within the \texttt{<answer>} block. A positive reward is given if the judge determines that the content is free of descriptions of irrelevant environmental information. This measure penalizes descriptions of the extraneous environment and reinforces the model's focus on the dynamic action itself.
\end{enumerate}

\section{Caption Quality Analysis}
\label{sec:Supplementary Experiment}
\subsection{Qualitative Caption Comparisons}
\label{sec:Caption Quality Supplementary Assessment}
\begin{figure*}[!t]
    \centering
    \begin{subfigure}[b]{0.47\textwidth}
        \includegraphics[width=\linewidth]{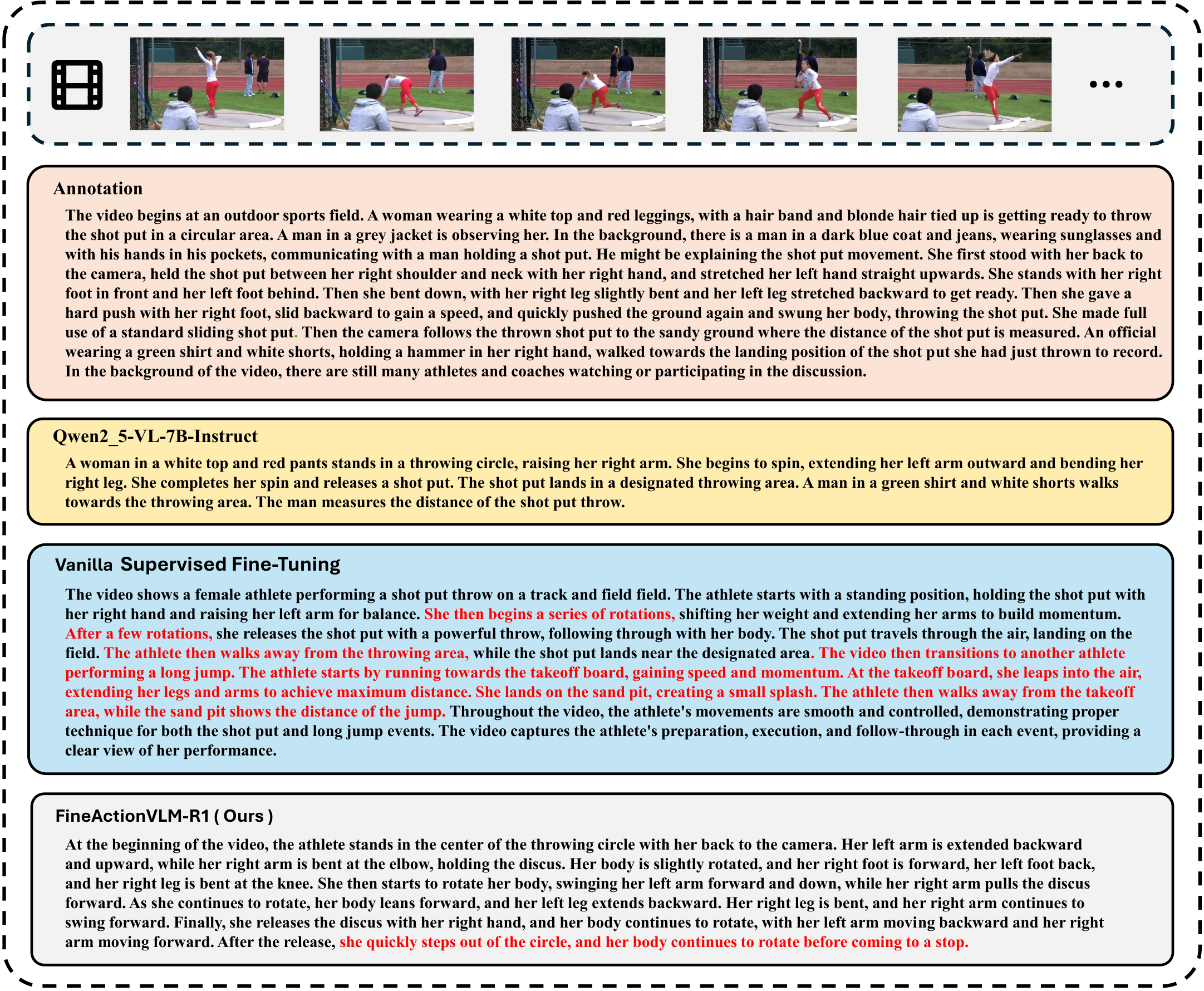}
        \caption{In-domain sample 2.}
        \label{fig:In domain2}
    \end{subfigure}
    \hfill
    \begin{subfigure}[b]{0.47\textwidth}
        \includegraphics[width=\linewidth]{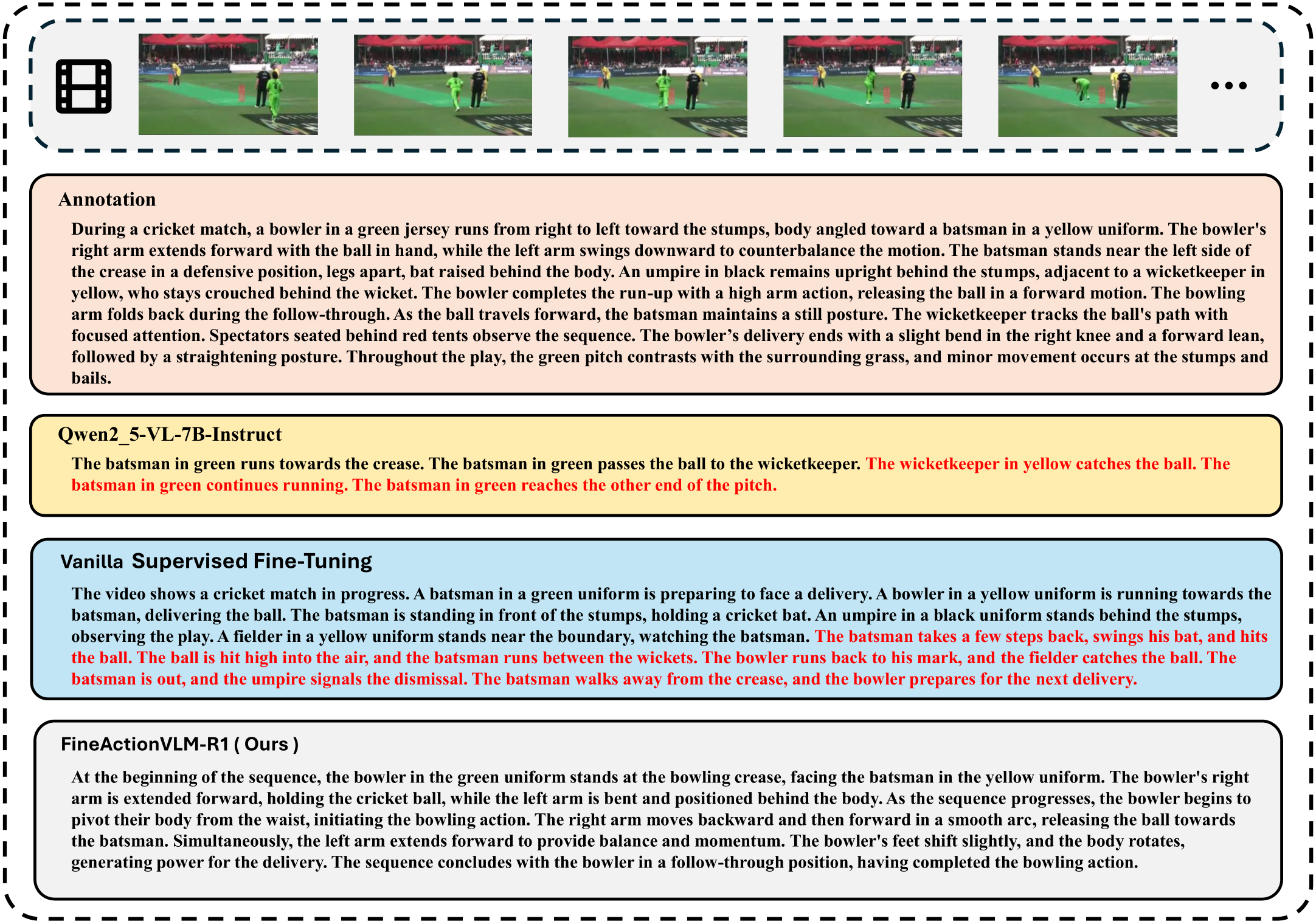}
        \caption{In-domain sample 3.}
        \label{fig:In domain3}
    \end{subfigure}

    \vspace{0.5em}

    \begin{subfigure}[b]{0.47\textwidth}
        \includegraphics[width=\linewidth]{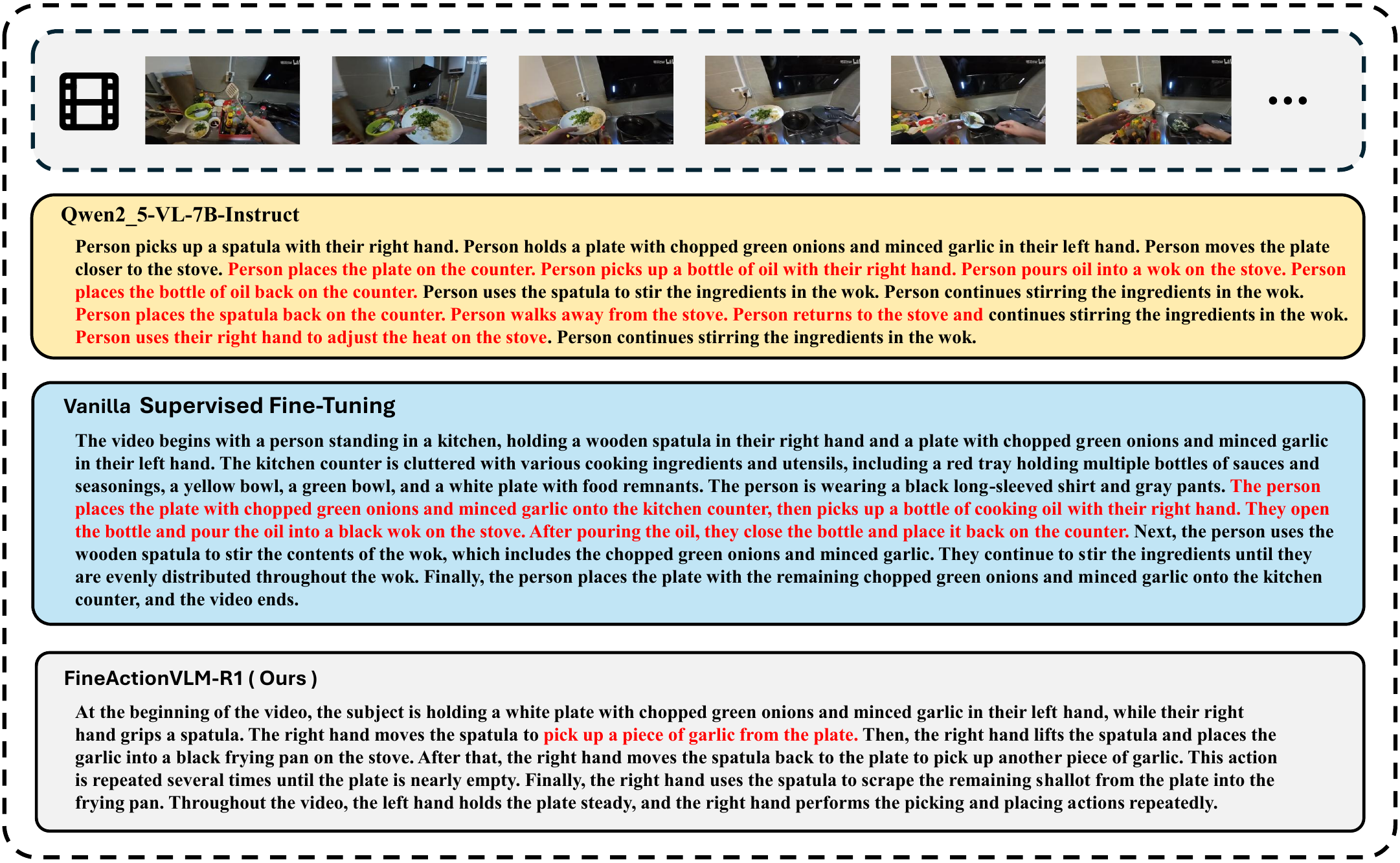}
        \caption{Out-of-domain sample 2.}
        \label{fig:Out of domain2}
    \end{subfigure}
    \hfill
    \begin{subfigure}[b]{0.47\textwidth}
        \includegraphics[width=\linewidth]{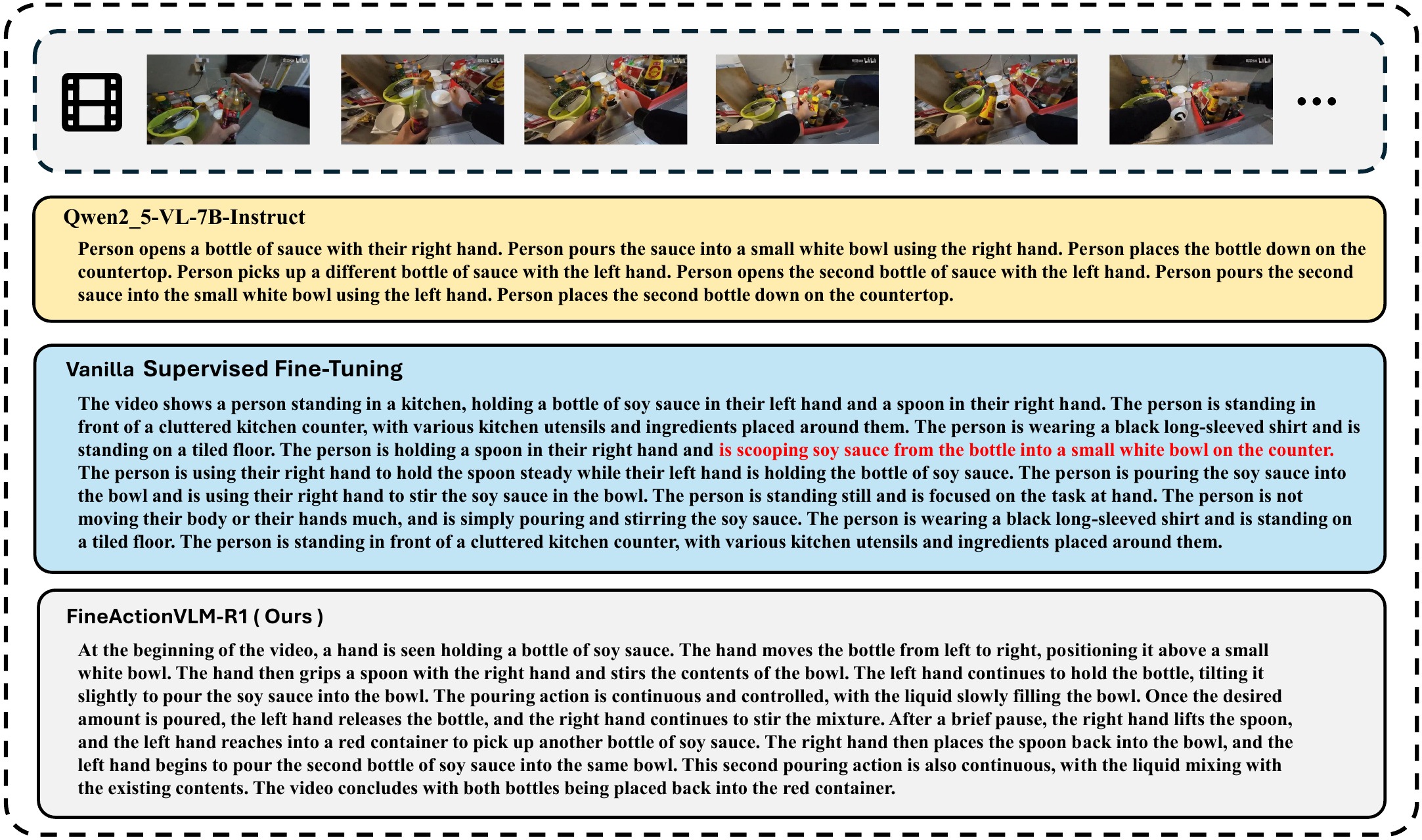}
        \caption{Out-of-domain sample 3.}
        \label{fig:Out of domain3}
    \end{subfigure}

    \caption{Caption Quality Supplementary Assessment.}
    \label{appendix_fig:cap}
\end{figure*}

In the in-domain evaluation on videos in Figure~\ref{fig:In domain2} and Figure~\ref{fig:In domain3}, the \textbf{Qwen2.5VL base model} provided generally coarse descriptions of the video content. For video in Figure~\ref{fig:In domain2}, the model was able to identify the overall motion pattern of the shot-put activity but failed to clearly disentangle and articulate the fine-grained motion details. In video in Figure~\ref{fig:In domain3}, the model only produced a global description of the athlete's movement in the stadium, accompanied by partial hallucinations that extended beyond the visual evidence.

The \textbf{vanilla SFT model} generated longer and more content-rich text compared with Qwen2.5VL, providing additional contextual details at the scene level. However, it did not show a substantial improvement in fine-grained action understanding. In video in Figure~\ref{fig:In domain2}, the model tended to hallucinate due to the presence of background objects such as a horizontal bar, incorrectly associating the scene with a long jump activity. Similarly, in video in Figure~\ref{fig:In domain3}, it generated several speculative descriptions about the athlete's movement based on perceptual inference rather than visible evidence.

The \textbf{FineActionVLM-R1 model} demonstrated a more structured understanding of the actions in video in Figure~\ref{fig:In domain2}. It accurately decomposed the athlete's movements in the shot-put sequence---from bending down for preparation to releasing the shot---providing a higher level of detail compared with the human-provided annotations. Although the motions of other participants were not mentioned, the main actor's actions were described with considerable granularity. In video in Figure~\ref{fig:In domain3}, FineActionVLM-R1 consistently focused on the primary athlete and precisely delineated the bowler's entire motion trajectory from the run-up to the ball release.

For the out-of-domain evaluation on videos in Figure~\ref{fig:Out of domain2} and Figure~\ref{fig:Out of domain3}, the \textbf{Qwen2.5VL base model} introduced substantial hallucinations in video in Figure~\ref{fig:Out of domain2}. The initial hallucinations arose from misinterpreting contextual elements such as soy sauce bottles on the kitchen counter, while the later ones appeared as imaginative extensions of the observed actions. In video in Figure~\ref{fig:Out of domain3}, no hallucinations were detected, but the model presented only coarse descriptions of the actors' activities, omitting significant motion details.

\FloatBarrier

The \textbf{vanilla SFT model} exhibited similar behavior on video in Figure~\ref{fig:Out of domain2}, with comparable levels of hallucination and fine-grained description as Qwen2.5VL, though it incorporated slightly more environmental context. On video in Figure~\ref{fig:Out of domain3}, the model provided a more complete narrative of the sauce-pouring process but still included minor hallucinatory elements.

In contrast, the \textbf{FineActionVLM-R1 model} generated hallucination-free descriptions for video in Figure~\ref{fig:Out of domain2}. While its fine-grained depiction was less detailed than that observed on in-domain videos, it clearly conveyed the overall procedural flow of the action. For video in Figure~\ref{fig:Out of domain3}, FineActionVLM-R1 demonstrated a precise understanding of both spatial and temporal dimensions, comprehensively describing the process of mixing sauces with clear and structured articulation of the corresponding actions.

\subsection{OOD Environment-Description Analysis}
\label{sec:original_ood_analysis}
To validate the efficacy of our proposed ``Attention Rectification'' strategy, we conducted an in-depth comparison between our model trained with this method (\textbf{FineActionVLM-R1-Dev1}) and a baseline model trained using only Vanilla SFT. Under identical conditions (temperature = 1, fps = 4), both models were evaluated on 11 different OOD videos. For each video, 25 responses were generated, resulting in a total of 275 samples per model. Using the DeepSeek-R1 evaluator, we analyzed the proportion of responses containing environmentally descriptive content that was irrelevant to human actions. The results show that 23.63\% of the Vanilla SFT model's responses contained such irrelevant environmental descriptions, whereas this proportion dropped to 15.63\% for FineActionVLM-R1-Dev1, an absolute reduction of 8.00 percentage points. This metric measures irrelevant environmental descriptions, rather than unsupported action claims, and is therefore indirect evidence. The 275 responses per model are repeated generations from 11 videos, not 275 independent videos. Direct human-rated action hallucination and action recall are reported separately in Appendix~\ref{sec:additional_ablation}.

\begin{figure*}[t]
    \centering
    \begin{subfigure}[b]{0.8\textwidth}
        \includegraphics[width=\textwidth]{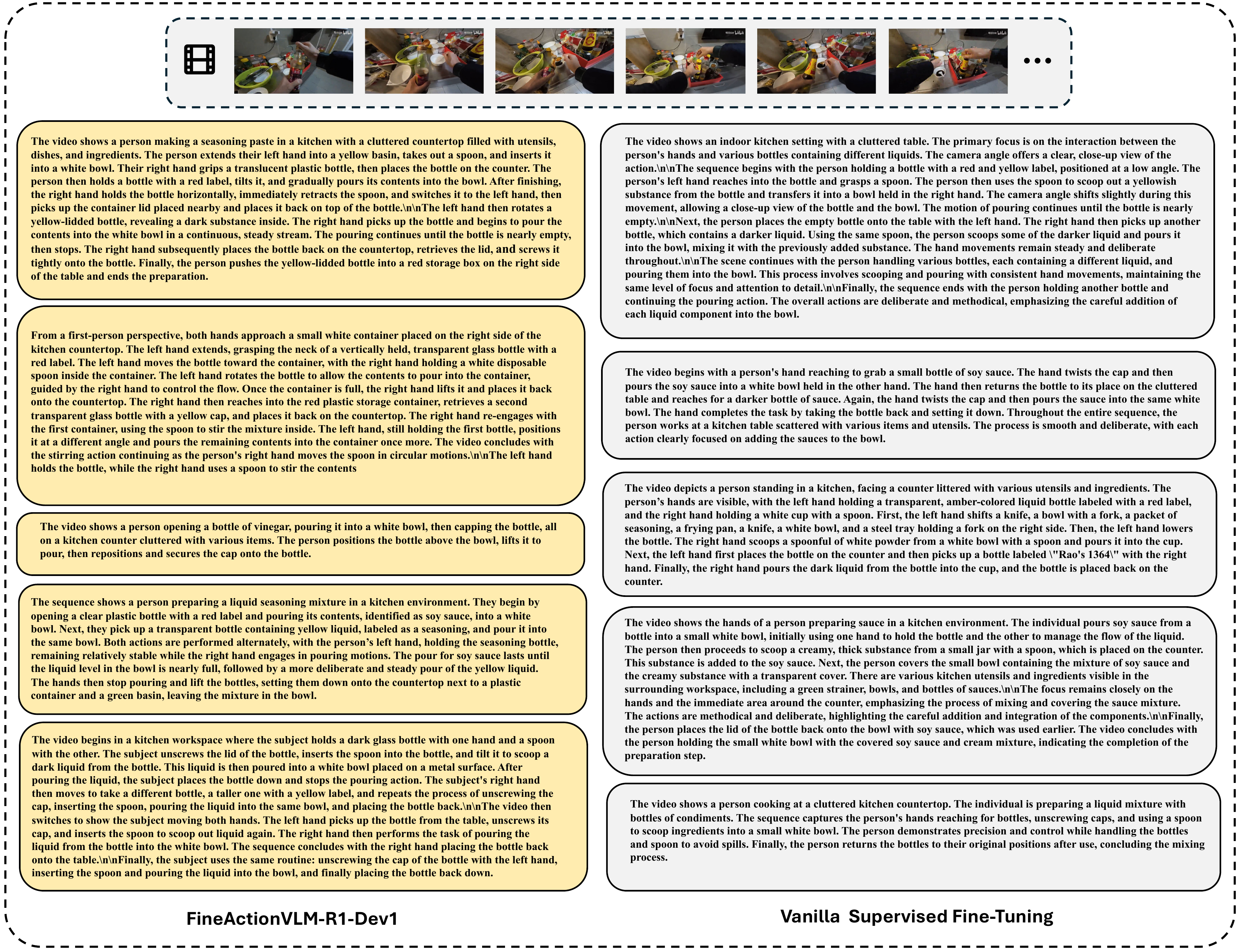}
    \end{subfigure}

    \vspace{5 pt}
    \begin{subfigure}[b]{0.8\textwidth}
        \includegraphics[width=\textwidth]{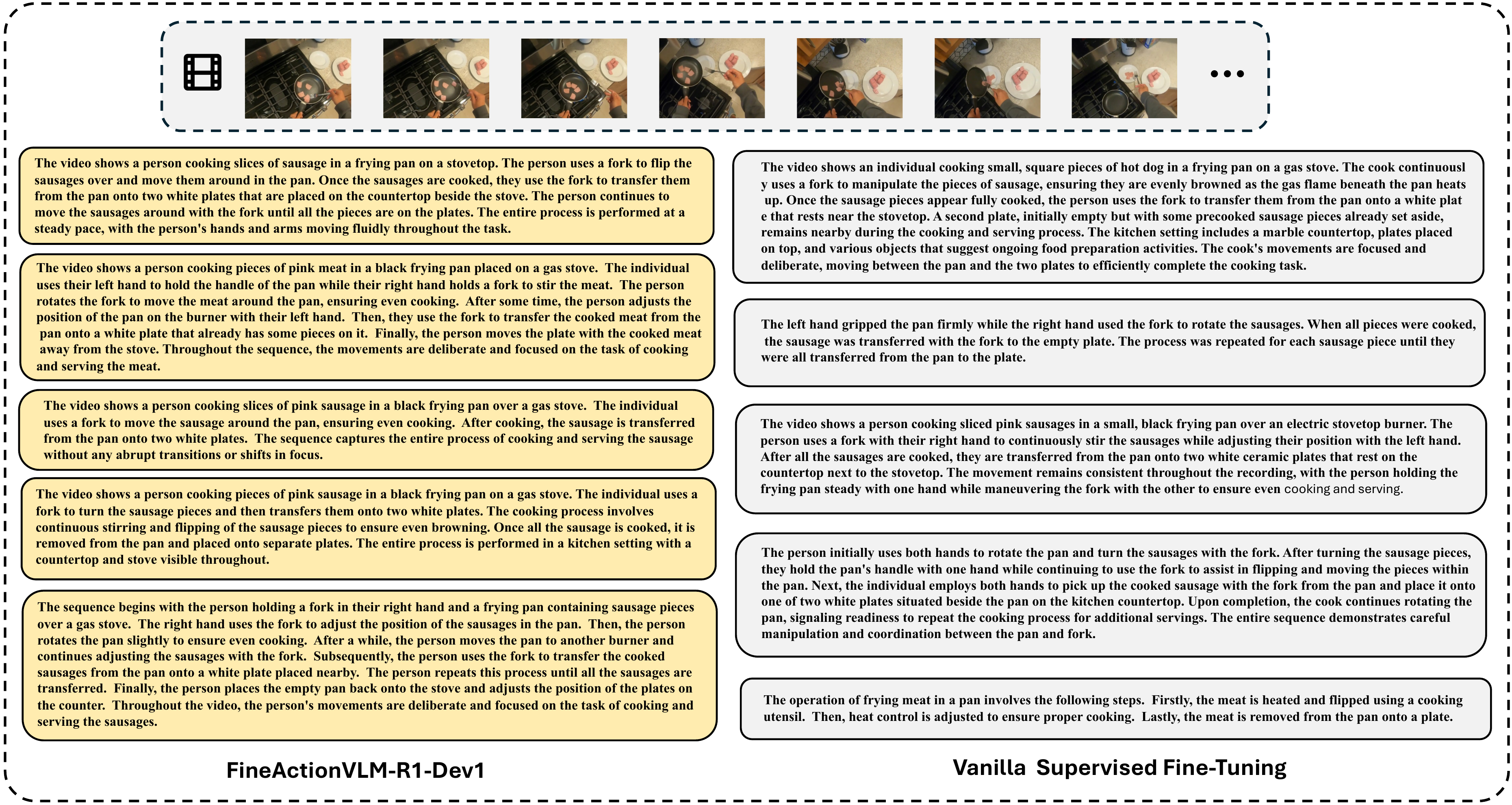}
    \end{subfigure}
   
    \caption{FineActionVLM-R1-Dev1 vs. Vanilla SFT}
    \label{fig:Dev1-vsft}
\end{figure*}
\FloatBarrier

When processing OOD videos, FineActionVLM-R1-Dev1 demonstrates significant advantages. As illustrated by the qualitative examples in Figures~\ref{fig:Dev1-vsft}, the descriptions generated by FineActionVLM-R1-Dev1 are consistently superior in overall quality to those from the baseline. Specifically, it is capable of capturing and articulating more fine-grained dynamic action details, its generated text is richer and more comprehensive, and the phenomenon of hallucination---generating content not present in the video---is effectively mitigated.

However, we also acknowledge that FineActionVLM-R1-Dev1 is not without its limitations. It still produces hallucinatory content in certain instances, and its generation quality can exhibit a degree of instability. We posit that these limitations may stem from two primary factors: the inherent capability ceiling of the foundation model used, and a lack of sufficient diversity in the training data, which constrains the model's generalization ability on certain video types. A noteworthy observation is that for videos that are either exceptionally simple or difficult, the performance gap between the two models is less pronounced. This further corroborates our hypothesis that the model's ultimate performance is constrained by both the training strategy and the fundamental dimensions of the base model and data.

Despite these challenges, the model trained with Attention Rectification provides a superior policy starting point for our subsequent optimization stages. Although its generated candidate descriptions may have occasional flaws, their overall quality is higher, laying a more solid foundation for our subsequent GRPO stage. This implies that GRPO can learn and optimize within a higher-quality and more promising candidate space, which is expected to lead to a higher performance ceiling. This finding also highlights future avenues for improvement: namely, that adopting a more powerful foundation model or further expanding the diversity of the training data could significantly enhance the model's overall performance.

\clearpage
\input{appendix_additional_experiments}

\end{document}

%% file: appendix_additional_experiments.tex
\section{Ablation Studies and Extended Evaluation}
\label{sec:additional_evaluation}

Complementing the caption analyses in Appendix~\ref{sec:Supplementary Experiment}, we evaluate human-rated action hallucination, isolate component contributions, and examine sampling sensitivity, task adaptation, token selection, and generalization. These experiments distinguish caption faithfulness from multiple-choice accuracy and examine the contributions of AED, standard GRPO, ACR, and Environmental Information Suppression (ES). Unless otherwise specified, Base denotes Qwen2.5-VL-7B-Instruct and Full or R1 denotes FineActionVLM-R1.

\subsection{Evaluation Protocol and Data Organization}
\label{sec:additional_protocol}

\paragraph{Human evaluation.}
The human evaluation covers 400 videos and involves five annotators. Each video is independently assessed by two annotators drawn from this pool, providing overlapping paired assessments across the annotation team. Model identities are blinded during evaluation. The reported inter-annotator agreement score is 0.82. Action hallucination (AH) is the percentage of captions containing actions unsupported by the video; action recall measures coverage of the ground-truth actions. We report both metrics because reducing unsupported claims is useful only if relevant action content is retained. AH directly evaluates action errors, while attributing a particular error to environmental cues requires additional controlled evidence.

\paragraph{Video sampling and source review.}
The dataset sampling policy uses a 1:1 ratio of first-person to third-person videos. Source-video deduplication involves 15 reviewers, with review within groups followed by cross-checking across groups to prevent duplicate original videos in the source pool. This source-review procedure is separate from the five-person human-evaluation protocol above. Multiple captions or questions can be derived from a single source video; source deduplication therefore does not imply that every derived annotation uses a different video.

\paragraph{HAU split scope.}
HAU-MCQ-Bench contains 753 test questions. The seen-video subset contains new questions about videos seen in HAU-MCQ-Train; the unseen-video subset contains both videos and questions unseen relative to HAU-MCQ-Train. Throughout this evaluation, ``unseen'' is defined relative to HAU-MCQ-Train.

\paragraph{Training and comparison structure.}
The three-stage recipe in the main-text \emph{Training Recipe} subsection contains an auxiliary reasoning-data stage and the two STARR optimization stages. First, GRPO on TASC-MCQ produces FineActionVLM-R1-Zero, which supplies distilled reasoning chains. Second, multi-task SFT starts from the base model, using fine-action captions, HAU-QA, distilled reasoning data, and AED construction where enabled, to obtain R1-Dev1. Third, GRPO starts from R1-Dev1 and uses captioning and multiple-choice tasks to obtain R1. Thus, R1-Zero supplies supervision for SFT; it is not the initialization checkpoint for that SFT stage.

Matched SFT is the supervised adaptation control. The component comparisons use matched supervision, with AED changing the construction of the SFT video input and ACR and ES varied within the GRPO stage. Base-to-SFT comparisons include both exposure to the task data and supervised optimization. Subsequent comparisons measure conditional component effects; they do not separately ablate every constituent dataset. Training hyperparameters and the shared task rewards are given in Appendix~\ref{sec:appendix_training}.

\paragraph{MotionBench aggregation.}
The Dev Avg column in Table~\ref{tab:experiment_results} is the overall accuracy pooled across questions. If task $k$ contains $n_k$ questions and has percentage accuracy $a_k$, the reported value is
\begin{equation}
    \mathrm{Avg}=\frac{\sum_{k=1}^{6} n_k a_k}{\sum_{k=1}^{6} n_k}.
    \label{eq:question_weighted_accuracy}
\end{equation}
It is therefore weighted by task size, rather than the unweighted mean of the six displayed task accuracies. This explains why directly averaging those columns does not reproduce the reported 54.83\% for Base or 57.84\% for R1. All differences below are expressed in percentage points (pp), unless explicitly stated otherwise.

\subsection{Component Ablations and Human-Rated Faithfulness}
\label{sec:additional_ablation}

Table~\ref{tab:additional_ablation} combines MotionBench accuracy with human-rated AH and action recall. We use G for standard GRPO, A for ACR, and E for ES. The A-only and E-only reward additions share the SFT+AED+G control; Full includes both ACR and ES.

\begin{table}[t]
    \centering
    \small
    \setlength{\tabcolsep}{4pt}
    \begin{tabular}{@{}lrrr@{}}
        \toprule
        Training & MB$\uparrow$ & AH$\downarrow$ & Recall$\uparrow$ \\
        \midrule
        Base & 54.83 & 29.78 & 66.41 \\
        Matched SFT & 55.27 & 28.76 & 70.23 \\
        SFT+AED & 55.83 & 25.24 & 71.91 \\
        SFT+G & 55.69 & 26.60 & 71.53 \\
        SFT+AED+G & 56.21 & 23.53 & 72.74 \\
        SFT+AED+G+A & 56.68 & 21.30 & 74.28 \\
        SFT+AED+G+E & 56.60 & 22.32 & 73.60 \\
        Full (+A+E) & \textbf{57.84} & \textbf{19.71} & \textbf{75.55} \\
        \bottomrule
    \end{tabular}
    \caption{Component ablations at 2 FPS. All values are percentages. MB is MotionBench Dev accuracy; AH and recall are human-evaluation metrics. G=GRPO, A=ACR, E=ES. Full adds both A and E to SFT+AED+G.}
    \label{tab:additional_ablation}
\end{table}

\paragraph{Data intervention and standard GRPO.}
Adding AED to matched SFT lowers AH from 28.76\% to 25.24\% (3.52 pp) and raises recall from 70.23\% to 71.91\% (1.68 pp). Standard GRPO without AED reduces AH to 26.60\%. Combining AED with GRPO reduces it further to 23.53\%, a 3.07 pp reduction relative to SFT+G. The corresponding MotionBench accuracy increases from 55.69\% to 56.21\%. These comparisons support a contribution from AED beyond supervised adaptation or standard GRPO alone.

\paragraph{ACR independently of ES.}
On top of SFT+AED+G, ACR lowers AH from 23.53\% to 21.30\% and raises recall from 72.74\% to 74.28\%. The AH difference is $-2.23$ pp with a paired 95\% confidence interval of $[-3.68,-0.78]$ pp. Because neither configuration includes ES, this improvement cannot be attributed solely to the environmental-suppression reward. Adding ES alone yields 22.32\% AH and 73.60\% recall; adding both rewards gives the best observed values, 19.71\% and 75.55\%.

\paragraph{Full-system comparison.}
Relative to matched SFT, Full improves MotionBench by 2.57 pp, reduces AH by 9.05 pp, and improves recall by 5.32 pp. The joint decrease in AH and increase in recall supports more faithful descriptions with greater action coverage, rather than a gain obtained simply by omitting actions. Table~\ref{tab:additional_ci} collects the paired intervals reported for the targeted comparisons. These intervals concern AH differences, not MotionBench accuracy or variation across training seeds.

\begin{table}[t]
    \centering
    \small
    \setlength{\tabcolsep}{3pt}
    \begin{tabularx}{\columnwidth}{@{}>{\raggedright\arraybackslash}Xrr@{}}
        \toprule
        Comparison (treatment $-$ control) & $\Delta$AH & 95\% CI \\
        \midrule
        SFT+AED $-$ matched SFT & $-3.52$ & $[-5.31,-1.73]$ \\
        Add ACR to SFT+AED+G, without ES & $-2.23$ & $[-3.68,-0.78]$ \\
        R1 $-$ Base at 4 FPS & $-11.11$ & $[-13.42,-8.80]$ \\
        \bottomrule
    \end{tabularx}
    \caption{Reported paired 95\% confidence intervals for AH differences, in percentage points. Negative differences favor the treatment.}
    \label{tab:additional_ci}
\end{table}

\subsection{Frame-Sampling Controls}
\label{sec:additional_sampling}

Sparse sampling can omit short actions. We therefore compare Base and R1 at 1, 2, and 4 FPS, using identical prompts, decoding settings, and visual-token limits across the comparison. Table~\ref{tab:additional_fps} reports both MotionBench accuracy and AH.

\begin{table}[t]
    \centering
    \small
    \begin{tabular}{@{}lrrrr@{}}
        \toprule
         & \multicolumn{2}{c}{MotionBench$\uparrow$} & \multicolumn{2}{c}{AH$\downarrow$} \\
        \cmidrule(lr){2-3}\cmidrule(l){4-5}
        FPS & Base & R1 & Base & R1 \\
        \midrule
        1 & 53.43 & 56.29 & 32.02 & 22.16 \\
        2 & 54.83 & 57.84 & 29.78 & 19.71 \\
        4 & 55.82 & 59.13 & 28.71 & 17.60 \\
        \bottomrule
    \end{tabular}
    \caption{Frame-sampling controls. All values are percentages. The advantage of R1 persists at every evaluated sampling rate.}
    \label{tab:additional_fps}
\end{table}

Denser sampling benefits both models. Increasing Base from 2 to 4 FPS improves MotionBench by 0.99 pp and lowers AH by 1.07 pp. At the same 4 FPS rate, however, R1 remains 3.31 pp higher on MotionBench and 11.11 pp lower in AH than Base, with the AH confidence interval shown in Table~\ref{tab:additional_ci}. The gains therefore persist under denser sampling; increasing frame rate alone does not explain the observed model difference under these evaluation settings.

\subsection{HAU-MCQ: Task Adaptation and Visual Input}
\label{sec:additional_hau}

Table~\ref{tab:additional_hau} compares Base, matched SFT, R1, and R1 without video input on HAU-MCQ-Bench. Video-input evaluations use 2 FPS. The benchmark contains 753 questions in total, with ``unseen'' defined as in Appendix~\ref{sec:additional_protocol}.

\begin{table}[t]
    \centering
    \small
    \begin{tabular}{@{}lrr@{}}
        \toprule
        Model/input & Overall$\uparrow$ & Unseen$\uparrow$ \\
        \midrule
        Base & 45.29 & 45.30 \\
        Matched SFT & 85.86 & 82.83 \\
        R1 & \textbf{91.50} & \textbf{90.31} \\
        R1, no video & 37.05 & 33.36 \\
        \bottomrule
    \end{tabular}
    \caption{HAU-MCQ accuracy (\%). Unseen denotes unseen videos and questions relative to HAU-MCQ-Train.}
    \label{tab:additional_hau}
\end{table}

Matched SFT improves overall accuracy by 40.57 pp over Base, accounting for much of the 46.21 pp Base-to-R1 gain. R1 improves over matched SFT by a further 5.64 pp overall and 7.48 pp on unseen videos. This separates the substantial effect of task adaptation from the subsequent improvement of the full training pipeline. Consequently, the headline comparison with larger models should be interpreted in light of differences in training exposure, rather than as a controlled comparison of model scale alone.

Removing video reduces R1 accuracy by 54.45 pp overall and 56.95 pp on the unseen subset. This provides evidence that visual input materially contributes to the result. It does not imply that all linguistic or question-format cues have been eliminated. HAU-MCQ and MotionBench assess different capabilities and involve different training exposure, so their gains need not have the same magnitude.

\subsection{Validation of Reference-Selected Visual Tokens}
\label{sec:additional_tokens}

ACR selects the top 25\% of visual tokens according to the reference model's attention to the ground-truth caption. We compare these selections with per-video random selections matched in token count, using annotated action evidence, including the temporal intervals in Fine-Action-Segment. Table~\ref{tab:additional_tokens} reports temporal and spatiotemporal overlap.

\begin{table}[t]
    \centering
    \small
    \setlength{\tabcolsep}{4pt}
    \begin{tabular}{@{}lrr@{}}
        \toprule
        Selection & Temporal & Spatiotemporal \\
        \midrule
        Matched random & 39.71 & 20.46 \\
        Reference top 25\% & \textbf{64.54} & \textbf{44.06} \\
        \bottomrule
    \end{tabular}
    \caption{Overlap with annotated action evidence (\%). Reference selection improves temporal overlap by 24.83 pp and spatiotemporal overlap by 23.60 pp.}
    \label{tab:additional_tokens}
\end{table}

Reference-selected tokens align more closely with annotated evidence than the matched random control. This validates the selection empirically, while allowing for residual reference-model bias. Overlap establishes association with annotated evidence; it does not establish that the selected tokens causally determine action predictions.

\paragraph{Interpretation of ACR.}
The attention value $a_i$ in the main-text \emph{Attention Consistency Reinforcement} formulation is averaged over generated text positions before $R_{\mathrm{var}}$ is computed across selected visual tokens. The variance term therefore encourages balanced caption-averaged coverage of those tokens. Two attention trajectories with identical averages receive the same variance reward even if their behavior across decoding steps differs. ACR promotes strong and balanced attention to the selected evidence; it does not directly measure or guarantee generation-time stability.

\subsection{Same-Scene, Different-Action Discrimination}
\label{sec:additional_pairs}

AED changes the clips surrounding an action segment while retaining the original scene and object cues within that segment. To examine action discrimination when the scene is shared, we additionally evaluate 100 same-scene/different-action video pairs. Table~\ref{tab:additional_pairs} reports correct-pair counts alongside AH from the separate human-caption evaluation.

\begin{table}[t]
    \centering
    \small
    \begin{tabular}{@{}lrr@{}}
        \toprule
        Training & Correct pairs/100 & AH (\%) \\
        \midrule
        Matched SFT & 54 & 28.76 \\
        SFT+AED & \textbf{71} & \textbf{25.24} \\
        \bottomrule
    \end{tabular}
    \caption{Same-scene action discrimination and separate human-rated AH. AH is not computed from the 100-pair task.}
    \label{tab:additional_pairs}
\end{table}

AED increases correct-pair counts from 54 to 71. Separately, its AH reduction relative to matched SFT is 3.52 pp, with a paired 95\% confidence interval of $[-5.31,-1.73]$ pp. Together, these results support improved action discrimination and caption faithfulness. The intervention targets correlations with surrounding context; it does not remove all scene or object information inside the action clip or establish complete scene--action disentanglement.

\subsection{General Video Understanding and Model Scale}
\label{sec:additional_generalization}

We evaluate general video understanding on Video-MME and test the training framework at the 3B scale within the Qwen2.5-VL family. Tables~\ref{tab:additional_videomme} and~\ref{tab:additional_3b} summarize the results.

\begin{table}[t]
    \centering
    \small
    \begin{tabular}{@{}lr@{}}
        \toprule
        Model (7B) & Video-MME accuracy (\%)$\uparrow$ \\
        \midrule
        Base & 56.00 \\
        Matched SFT & 56.45 \\
        Full & \textbf{57.77} \\
        \bottomrule
    \end{tabular}
    \caption{General video-understanding performance. Full improves over matched SFT by 1.32 pp.}
    \label{tab:additional_videomme}
\end{table}

\begin{table}[t]
    \centering
    \small
    \begin{tabular}{@{}lrr@{}}
        \toprule
        Qwen2.5-VL-3B & MotionBench$\uparrow$ & AH$\downarrow$ \\
        \midrule
        Base & 51.49 & 31.22 \\
        Full training & \textbf{53.40} & \textbf{23.89} \\
        \bottomrule
    \end{tabular}
    \caption{Transfer to the 3B backbone. All values are percentages.}
    \label{tab:additional_3b}
\end{table}

On Video-MME, Full gains 1.77 pp over Base and 1.32 pp over matched SFT, extending the observed improvement beyond the fine-grained benchmarks. At 3B scale, full training improves MotionBench by 1.91 pp and reduces AH by 7.33 pp. These results support transfer across the evaluated 3B and 7B scales within Qwen2.5-VL. They do not yet establish transfer across distinct backbone architectures.

\subsection{Discussion of Caption Faithfulness}
\label{sec:additional_synthesis}

Appendix~\ref{sec:original_ood_analysis} evaluates 11 OOD videos at 4 FPS, with 25 generated responses per video and model, yielding 275 responses per model. Its DeepSeek-R1-based score measures irrelevant environmental descriptions: 23.63\% for Vanilla SFT and 15.63\% for R1-Dev1, an absolute decrease of 8.00 pp. This score is distinct from human-rated AH. Repeated generations from 11 videos do not constitute 275 independent videos, and the environmental-description score is indirect evidence rather than a measurement of unsupported actions. The human evaluation supplies the direct caption-faithfulness measurement, while recall checks action coverage.

Taken together, these experiments show that AED and ACR provide conditional improvements beyond the matched SFT and standard GRPO controls, including an ACR benefit without ES. These gains coexist with higher action recall, persist under denser frame sampling, and extend to Video-MME and a smaller model within the same family. Token overlap and same-scene discrimination provide complementary evidence of improved use of action-related information. The supported conclusion is improved action faithfulness and discrimination under the evaluated settings, with the scope of each mechanism grounded in its actual objective and controls.